\documentclass{article} 
\usepackage{iclr2027_conference,times}

\usepackage{amsmath,amsfonts,bm}

\def\eqref#1{equation~\ref{#1}}

\def\1{\bm{1}}

\DeclareMathAlphabet{\mathsfit}{\encodingdefault}{\sfdefault}{m}{sl}
\SetMathAlphabet{\mathsfit}{bold}{\encodingdefault}{\sfdefault}{bx}{n}

\usepackage{hyperref}
\usepackage{url}
\usepackage{booktabs}
\usepackage{graphicx}
\usepackage{wrapfig}
\usepackage{placeins}
\usepackage{tikz}
\usetikzlibrary{arrows.meta,calc}

\title{A Persistent State for Auditable Mixture-of-Experts Routing}

\author{Abdurrahman Javat,~ Allan Kazakov  \\
Department of Artificial Intelligence\\
Bahçeşehir University\\
Istanbul, Türkiye \\
\texttt{\{abdurrahman.javat,allan.kazakov\}@bahcesehir.edu.tr} \\
}

\iclrfinalcopy 
\begin{document}

\maketitle

\begin{abstract}

Mixture-of-Experts (MoE) models repeatedly route tokens to sparse subsets of
experts, but conventional routers expose no routing-specific record of how
cross-layer influences accumulate. We introduce Scratchpad-Augmented
Mixture-of-Experts (SA-MoE), which gives each router access to a
low-dimensional persistent state that is not provided to the experts. Learned
layerwise writes update this state, and their realized post-update changes
exactly decompose the state-mediated contribution to any later routing margin,
forming a routing ledger.
Across sparsely upcycled SmolLM2- and Gemma-based models and three independent
training seeds per architecture, this pathway adds less than 1\% analytical
forward compute and is strongly used by trained routers: local removal of its
router contribution changes the selected Top-2 expert set in 87.6\% and
69.9\% of decisions, respectively. Relative to a matched latest-write-only
control, persistent accumulation increases long-horizon future-routing
accessibility by 19.4 and 12.2 percentage points, with positive effects in
every seed. More than 90\% of absolute ledger contribution comes from
non-recent writes in both families, and full-forward suppression of
ledger-selected writes changes later routing and output distributions.
The ledger is an exact provenance object for the persistent-state pathway, not
a complete causal explanation of routing. Sensitivity-aware scores better predict full-forward intervention effects, and post-hoc methods recover related cross-layer
attribution without architectural modification. SA-MoE instead makes one
routing-specific computational history explicit and directly inspectable within
the model's natural forward computation.

\end{abstract}

\section{Introduction}
\label{sec:introduction}

Mixture-of-Experts (MoE) Transformers increase model capacity while keeping
per-token computation sparse by routing each token to a small subset of experts
\citep{shazeer2017outrageously,lepikhin2021gshard,fedus2022switch}.
Because routing is repeated throughout the network, understanding why a later
router favors one expert over another requires reasoning about influences that
may accumulate across depth. Standard MoEs expose no routing-specific record of
that history: routing is computed from the same general hidden stream used by
the rest of the model. Cross-layer routing influences can be reconstructed
post hoc, and prior architectures have introduced explicit memory between
routers, but these approaches do not make write-level routing provenance a
first-class object of the forward computation.

We introduce Scratchpad-Augmented Mixture-of-Experts (SA-MoE), which gives
each router access to a low-dimensional persistent state that is not provided
to the experts. A learned write updates this state after each MoE layer.
The realized post-update state changes form a routing ledger: for any later
router margin, the state-mediated contribution decomposes exactly into
contributions from earlier realized writes. We do not claim that telescoping a
recorded recurrent trajectory is unique to SA-MoE; the same algebra can be
applied to other recorded recurrent states. The architectural distinction is
that SA-MoE defines learned write events as the native provenance object, with
a direct source-level intervention semantics.

We test whether this provenance object is useful rather than merely
algebraically available. Across sparsely upcycled SmolLM2- and Gemma-based
models and three independent training seeds per architecture, trained routers
strongly use the state pathway. Relative to a matched latest-write-only control,
persistent accumulation increases long-horizon future-routing accessibility by
19.4 and 12.2 percentage points, respectively, and more than 90\% of the
ledger's absolute contribution comes from non-recent writes in both families.
Ledger-selected writes also affect later routing and output distributions under
full-forward suppression. At the same time, sensitivity-aware
gradient$\times$write can provide stronger intervention rankings, and post-hoc
attribution on matched ordinary MoEs recovers related cross-layer influences
without architectural modification.

Our contribution is therefore an MoE routing design in which one
cross-layer pathway carries an explicit, inspectable computational history.
SA-MoE couples a compact routing-specific state with write-level provenance,
retrospective routing queries, and direct intervention on the recorded source
events. This provides a complementary operating point to recurrent routing
memory, hidden-stream probing, and post-hoc attribution rather than replacing
them.

\section{Background and Related Work}
\label{sec:related}

\subsection{Sparse MoE Routing and Cross-Layer Memory}

Sparse MoEs activate only a small subset of experts for each token
\citep{shazeer2017outrageously,lepikhin2021gshard,fedus2022switch}, making
routing a repeated discrete allocation decision across model depth.
Cross-layer routing memory itself is not new. RMoE \citep{qiu2025layerwise} uses a
GRU-based recurrent router that carries routing-specific information between
successive MoE layers. Moreover, if a recurrent trajectory $h_\ell$ is
recorded, any fixed linear target readout admits the identity
\[
W_L h_L
=
W_L h_0
+
\sum_{\ell<L} W_L(h_{\ell+1}-h_\ell),
\]
including when the recurrence generating $h_\ell$ is nonlinear. Exact
telescoping of realized state differences is therefore not unique to SA-MoE.

SA-MoE instead makes the update itself an explicit audit interface. A learned
write proposal produces a realized post-clamp write that is stored as the
source-level provenance object and can subsequently be queried or suppressed at
its source. Our latest-write-only control tests persistence against replacement
while retaining this state-and-write pathway. Memory-Aware Routing
\citep{hou2026memory} addresses a different form of memory, maintaining
expert-level routing preferences to encourage stable specialization rather than
a per-token state propagated across model depth.

\subsection{Architecture-Level MoE Interpretability}

Other work modifies MoE computation to make internal structure more
interpretable. MoE-X \citep{yang2025mixture} relates experts to a wide sparse MLP
and encourages sparse expert computation, while MONET \citep{park2025monet}
incorporates sparse dictionary-learning ideas into MoE training to promote
more interpretable expert structure. These approaches target semantic or
feature-level interpretability. SA-MoE does not require experts or individual
state coordinates to have semantic meanings; its audit object is instead the
computational provenance of one routing-specific pathway.

\subsection{Post-Hoc Attribution and Causal Validation}

A complementary line of work reconstructs routing influences after training.
\citet{li2025decoding} propose cross-level knowledge attribution for ordinary
MoE computation, while \citet{li2026understanding} recursively decompose later
router inputs into contributions from earlier model components and identify
cross-layer routing influences that can persist across depth.
We therefore do not claim that cross-layer routing attribution requires a
dedicated routing state. The distinction is where the audit object originates:
these post-hoc methods reconstruct contributions from ordinary model
computation, whereas SA-MoE places a routing-specific state and explicit write
events inside the natural forward computation. We evaluate the method of
\citet{li2026understanding} directly on matched Ordinary MoEs rather than treating
it only as related work.

Attribution also differs from causal sensitivity. Activation replacement,
ablation, and patching are widely used to test whether identified components
affect model behavior \citep{meng2022locating,conmy2023automated}, including
recent analyses that intervene directly on MoE routing
\citep{bandarkar2026multilingual,xu2026routingdistraction}.
Our full-forward intervention follows the same general logic: a selected
realized write is removed at its source and all subsequent computation is
allowed to recompute. Gradient-based rankings provide a complementary
sensitivity-aware comparison. Accordingly, our experiments distinguish exact
forward provenance, post-hoc attribution, and intervention sensitivity rather
than treating them as the same objective.

\section{Scratchpad-Augmented Mixture-of-Experts}
\label{sec:method}

We study a sparse MoE architecture in which routers receive a
routing-specific cross-layer state, while the experts themselves do not.
At MoE layer $\ell$, the ordinary token representation is
$x_\ell \in \mathbb{R}^{d_x}$ and the routing state is
$s_\ell \in \mathbb{R}^{d_s}$.
The router computes expert logits
\begin{equation}
z_\ell = W_x^\ell x_\ell + W_s^\ell s_\ell + b^\ell,
\label{eq:router_logits}
\end{equation}
after which Top-$k$ routing selects the routed experts.
As in an ordinary sparse MoE, the selected routed experts and a shared
expert contribute to the next hidden representation $x_{\ell+1}$.
The key difference is that only the router reads $s_\ell$;
the experts operate on $x_\ell$ alone.

A learned write head updates the routing state after each MoE layer.
Let $u_\ell$ denote the write proposal produced from the layer's shared
representation.
In the persistent architecture, the state evolves as
\begin{equation}
s_{\ell+1} = \operatorname{clip}(s_\ell + u_\ell,\,-5,\,5).
\label{eq:persistent_update}
\end{equation}
We define the realized write
\begin{equation}
\Delta s_\ell \equiv s_{\ell+1} - s_\ell,
\label{eq:realized_write}
\end{equation}
which is the post-clamp state change actually retained by the model.

\paragraph{Exact routing ledger.}
Consider a target router at layer $L$ and an audited margin between
experts $a$ and $b$ arising from the state pathway alone:
\begin{equation}
m_{s,L}^{a,b}
=
\left(W_{s,a}^L - W_{s,b}^L\right)^\top s_L.
\label{eq:state_margin}
\end{equation}
Because the persistent state is the accumulated sum of realized writes,
\begin{equation}
s_L = \sum_{\ell < L} \Delta s_\ell,
\label{eq:state_sum}
\end{equation}
the same margin decomposes exactly as
\begin{equation}
m_{s,L}^{a,b}
=
\sum_{\ell < L}
c_{\ell \to L}^{a,b},
\qquad
c_{\ell \to L}^{a,b}
=
\left(W_{s,a}^L - W_{s,b}^L\right)^\top \Delta s_\ell.
\label{eq:ledger}
\end{equation}
We call this exact write-level decomposition the
\emph{routing ledger}.
It is exact for the persistent-state contribution to the audited router
margin.
Other influences on routing, including those mediated through the
ordinary hidden stream $x$, remain outside this decomposition.

\paragraph{Controls.}
We compare Persistent SA-MoE against two matched controls:

\textbf{Ordinary MoE} removes the routing-state pathway entirely, so
routing depends only on $x_\ell$.

\textbf{Latest-write-only} keeps the same state width and write-head
capacity, but removes cross-layer accumulation by replacing the state
rather than accumulating it.
Thus the router at layer $\ell+1$ receives only the most recently
retained write-derived state.
Because this recurrence is a replacement rather than an accumulation,
its native audit object is not a telescoping sum of earlier state
differences.
Instead, its direct native attribution is structural: the currently
retained state is attributed to the latest retained write.

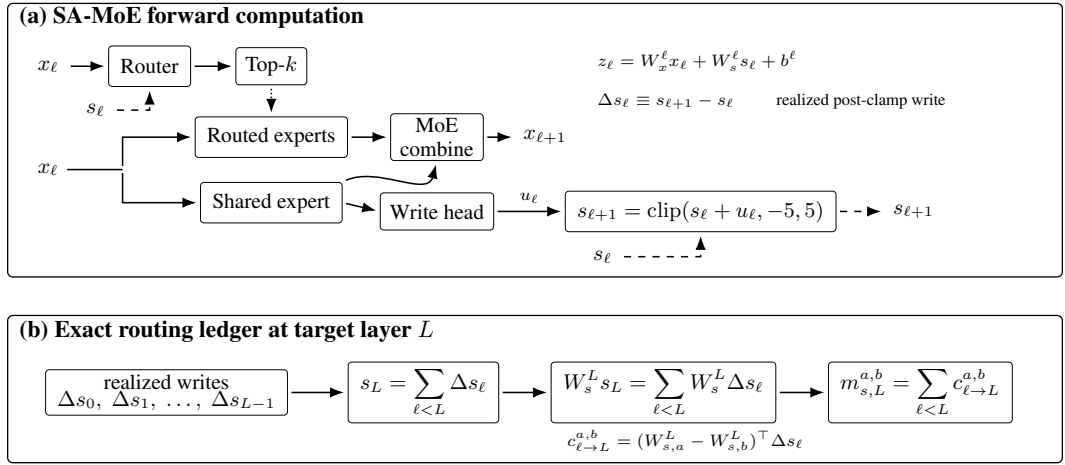
\begin{figure}[t]
\centering
\resizebox{\linewidth}{!}{%
\begin{tikzpicture}[
  font=\footnotesize,
  >=Latex,
  box/.style={
    draw,
    rounded corners=1.5pt,
    align=center,
    minimum height=6mm,
    inner xsep=5pt,
    inner ysep=2pt
  },
  smallbox/.style={
    draw,
    rounded corners=1.5pt,
    align=center,
    minimum height=5.5mm,
    inner xsep=4pt,
    inner ysep=1.8pt
  },
  flow/.style={
    ->,
    line width=.65pt,
    shorten >=1.5pt,
    shorten <=1.5pt
  },
  control/.style={
    ->,
    densely dotted,
    line width=.65pt,
    shorten >=1.5pt,
    shorten <=1.5pt
  },
  state/.style={
    ->,
    dashed,
    line width=.65pt,
    shorten >=1.5pt,
    shorten <=1.5pt
  },
  frame/.style={
    draw,
    rounded corners=2.2pt,
    line width=.55pt
  }
]


\node[font=\bfseries,anchor=west]
  at (0,1.42)
  {(a) SA-MoE forward computation};


\node (xrouter)
  at (0.55,0.70)
  {$x_\ell$};

\node[box] (router)
  at (2.05,0.70)
  {Router};

\node[smallbox] (topk)
  at (3.85,0.70)
  {Top-$k$};

\draw[flow]
  (xrouter.east) -- (router.west);

\draw[flow]
  (router.east) -- (topk.west);

\node (srouter)
  at (1.25,0.05)
  {$s_\ell$};

\draw[state]
  (srouter.east) -| (router.south);


\node (xexpert)
  at (0.55,-0.83)
  {$x_\ell$};

\coordinate (xsplit)
  at (1.65,-0.83);

\node[box] (routed)
  at (3.85,-0.35)
  {Routed experts};

\node[box] (shared)
  at (3.85,-1.32)
  {Shared expert};

\draw[line width=.65pt]
  (xexpert.east) -- (xsplit);

\draw[flow]
  (xsplit) |- (routed.west);

\draw[flow]
  (xsplit) |- (shared.west);

\draw[control]
  (topk.south) -- (routed.north);


\node[smallbox] (combine)
  at (6.30,-0.35)
  {MoE\\combine};

\node (xnext)
  at (7.90,-0.35)
  {$x_{\ell+1}$};

\draw[flow]
  (routed.east) -- (combine.west);

\draw[flow]
  (shared.north east)
  to[out=25,in=-105]
  (combine.south);

\draw[flow]
  (combine.east) -- (xnext.west);


\node[smallbox] (write)
  at (6.30,-1.45)
  {Write head};

\node[box] (update)
  at (10.20,-1.45)
  {$
    s_{\ell+1}
    =
    \operatorname{clip}
    (s_\ell+u_\ell,-5,5)
   $};

\node (snext)
  at (13.35,-1.45)
  {$s_{\ell+1}$};

\draw[flow]
  (shared.east) -- (write.west);

\draw[flow]
  (write.east)
  --
  node[above,font=\scriptsize] {$u_\ell$}
  (update.west);

\draw[state]
  (update.east) -- (snext.west);

\node (supdate)
  at (8.75,-2.12)
  {$s_\ell$};

\draw[state]
  (supdate.east) -| (update.south);


\node[
  font=\scriptsize,
  anchor=west
] at (8.55,0.78)
  {$
    z_\ell
    =
    W_x^\ell x_\ell
    +
    W_s^\ell s_\ell
    +
    b^\ell
   $};

\node[
  font=\scriptsize,
  anchor=west
] at (8.55,0.18)
  {$
    \Delta s_\ell
    \equiv
    s_{\ell+1}-s_\ell
   $};

\node[
  font=\scriptsize,
  anchor=west
] at (11.20,0.18)
  {realized post-clamp write};

\draw[frame]
  (-.05,-2.42)
  rectangle
  (15.55,1.63);


\begin{scope}[yshift=-4.15cm]

\node[
  font=\bfseries,
  anchor=west
] at (0,0.92)
  {(b) Exact routing ledger at target layer $L$};

\node[box] (writes)
  at (2.30,0.02)
  {realized writes\\[-1mm]
   $\Delta s_0,\;
    \Delta s_1,\;
    \ldots,\;
    \Delta s_{L-1}$};

\node[box] (sum)
  at (6.10,0.02)
  {$
    s_L
    =
    \displaystyle\sum_{\ell<L}
    \Delta s_\ell
   $};

\node[box] (proj)
  at (9.65,0.02)
  {$
    W_s^L s_L
    =
    \displaystyle\sum_{\ell<L}
    W_s^L\Delta s_\ell
   $};

\node[box] (margin)
  at (13.45,0.02)
  {$
    m_{s,L}^{a,b}
    =
    \displaystyle\sum_{\ell<L}
    c_{\ell\to L}^{a,b}
   $};

\draw[flow]
  (writes.east) -- (sum.west);

\draw[flow]
  (sum.east) -- (proj.west);

\draw[flow]
  (proj.east) -- (margin.west);

\node[
  font=\scriptsize,
  anchor=west
] at (8.10,-0.70)
  {$
    c_{\ell\to L}^{a,b}
    =
    (W_{s,a}^L-W_{s,b}^L)^\top
    \Delta s_\ell
   $};

\draw[frame]
  (-.05,-1.02)
  rectangle
  (15.55,1.16);

\end{scope}

\end{tikzpicture}%
}

\caption{
\textbf{SA-MoE and the routing ledger.}
\textbf{(a)} Each router reads both the ordinary hidden representation
$x_\ell$ and a routing-specific persistent state $s_\ell$, while the
experts read only $x_\ell$.
A write head produces a realized state update after each MoE layer.
\textbf{(b)} Because the persistent state is the sum of realized writes,
the state-mediated contribution to a later routing margin decomposes
exactly into write-level contributions.
}
\label{fig:architecture-ledger}
\end{figure}

Figure~\ref{fig:architecture-ledger} summarizes the architecture and the
resulting exact ledger for the persistent pathway.

\section{Experimental Setup}
\label{sec:setup}

\paragraph{Model families and training.}
We study two sparsely upcycled \citep{drop-upcycling, komatsuzaki2023sparse} model families based on
SmolLM2-135M \citep{benallal2025smollm2} and Gemma-270M \citep{gemmateam2025gemma3}.
For each family, we train three matched sparse architectures:
an Ordinary MoE baseline, a latest-write-only control, and Persistent
SA-MoE.
All three variants use the same sparse backbone, with 16 routed experts,
one shared expert, Top-2 routing, and matched continued-pretraining
budgets of approximately 30B packed input tokens.
The two state-augmented variants use a 128-dimensional routing state.
Training data are sampled from FineWeb-Edu \citep{penedo2024fineweb}, The Stack Dedup Python \citep{thestack}, and
FineMath \citep{benallal2025smollm2} with probabilities $0.50$, $0.25$, and $0.25$.
We train three independent seeds (42, 43, and 44) for each
architecture in each family.
Full optimization, initialization, and implementation details are given
in Appendix~\ref{app:training}.

\paragraph{Capability evaluation.}
Before evaluating auditability, we first test whether adding the
routing-state pathway materially changes model quality.
We report validation perplexity and a four-task zero-shot downstream
suite consisting of HellaSwag \citep{zellers2019hellaswag}, ARC-Easy \citep{clark2018arc}, PIQA \citep{bisk2020piqa}, and WinoGrande \citep{sakaguchi2020winogrande}.
For the downstream suite, we report the unweighted macro average of the
task primary metrics.
We evaluate family-level capability preservation using frozen
non-inferiority thresholds:
at most a $1\%$ relative increase in perplexity and at most a
$1$ percentage point decrease in downstream macro score relative to the
matched Ordinary MoE.
Appendix~\ref{app:evaluation} gives the exact prompts, scoring rules,
and task-specific metrics.

\paragraph{State use and persistence.}
We first ask whether trained routers actually use the routing state.
For this analysis, we locally intervene on the state contribution to a
router while holding the ordinary hidden representation fixed, and
measure the resulting change in the selected Top-2 expert set.
We then compare Persistent SA-MoE against the latest-write-only control
to test whether cross-layer accumulation provides additional value
beyond a same-capacity non-persistent state pathway.

\paragraph{Future-routing accessibility.}
To test whether the routing state compactly exposes future-routing
information, we train probes to predict a later router's Top-1 expert
from either the native routing state or representations derived from the
ordinary hidden stream.
The main text compares three representation families:
the native 128-dimensional routing state,
a full-hidden linear probe,
and a supervised 128-dimensional hidden subspace.
We evaluate probe-training budgets of
$10^4$, $10^5$, and $10^6$ examples.
Appendix~\ref{app:probe_fitting} gives the full probe grids, fit
selection rules, and supporting PCA and random-projection controls.

\paragraph{Attribution and intervention protocol.}
Our confirmatory intervention study uses 24 fixed final-evaluation
sequences per model family, spanning prose, code, and mathematics, and
96 target routing decisions per family.
For each target decision at layer $L$, we audit the Top-2 membership
boundary between the clean second-selected expert $e_2$ and the clean
best-excluded expert $e_3$.
For Persistent SA-MoE, the ledger score of source layer $\ell < L$ is
the absolute write contribution
$\lvert c_{\ell \to L}^{(2,3)} \rvert$ from
Eq.~\ref{eq:ledger}.
We compare this ranking with recency, largest-write-norm, and
gradient$\times$write baselines.
We also apply the post-hoc cross-layer attribution method of \citet{li2026understanding} to matched Ordinary MoEs.
Because SA-MoE writes and Ordinary-MoE residual components are distinct
audit objects, we evaluate attribution quality within each object and do
not compare raw intervention magnitudes across architectures.
For latest-write-only, the native direct score is defined over its
structural latest-state audit object rather than over a telescoping sum
of earlier state differences.

We then perform full-forward source suppression.
In Persistent SA-MoE, suppressing a source write restores the
post-source state to its pre-write value and allows the rest of the
model to recompute naturally.
In latest-write-only, suppression restores the previous retained state.
We measure the resulting absolute change in the audited target margin,
the divergence of later Top-2 routing decisions, and the KL divergence
between the clean and intervened output distributions at the selected
token.
Appendix~\ref{app:attribution} gives the full deterministic
selection procedure, ranking definitions, and control rules.

\paragraph{Uncertainty.}
For architecture-level claims, the training seed is the replication
unit.
We form paired seed-level contrasts and report three-seed means with
95\% Student-$t$ intervals ($n=3$, $2$ degrees of freedom).
For within-checkpoint intervention quantities, uncertainty is estimated
with paired sequence-level bootstrap resamples; these bootstrap samples
are not treated as additional training replicates.

\begin{table*}[t]
\centering
\caption{
\textbf{Model quality and architectural cost across three training seeds.}
Quality columns are paired mean changes relative to Ordinary MoE;
brackets give 95\% Student-$t$ intervals over training seeds ($n=3$).
Macro is the unweighted mean of HellaSwag, ARC-Easy, PIQA, and
WinoGrande primary metrics.
}
\label{tab:quality_cost}
\scriptsize
\setlength{\tabcolsep}{5.0pt}
\begin{tabular}{llrrrrr}
\toprule
Family & Architecture &
Params & Active/token & FLOPs/token &
$\Delta$PPL (\%) & $\Delta$Macro (pp) \\
\midrule

SmolLM2
& Ordinary MoE
& 1.409B & 294.0M & 729.6M
& Ref. & Ref. \\

& Latest-write-only
& 1.411B & 296.3M & 734.1M
& $+0.52$ [$+0.18,+0.86$]
& $-0.02$ [$-0.48,+0.43$] \\

& Persistent
& 1.411B & 296.3M & 734.1M
& $+0.19$ [$+0.12,+0.25$]
& $-0.05$ [$-0.14,+0.04$] \\

\midrule

Gemma
& Ordinary MoE
& 1.401B & 409.8M & 876.2M
& Ref. & Ref. \\

& Latest-write-only
& 1.402B & 411.4M & 879.2M
& $+0.30$ [$-0.18,+0.79$]
& $-0.49$ [$-1.31,+0.33$] \\

& Persistent
& 1.402B & 411.4M & 879.2M
& $-0.06$ [$-0.45,+0.32$]
& $-0.16$ [$-1.66,+1.35$] \\

\bottomrule
\end{tabular}
\end{table*}

\section{Results}
\subsection{Model Quality and Architectural Cost}
\label{sec:results_quality}

We first ask whether adding the routing-state pathway materially changes
model quality or architectural cost.
Table~\ref{tab:quality_cost} reports paired architecture contrasts across
the three training seeds.
Persistent SA-MoE adds less than $1\%$ active parameters and analytical
forward FLOPs per token in both model families.

On SmolLM2, Persistent changes validation perplexity by
$+0.19\%$ (95\% CI $[+0.12,+0.25]\%$) and the four-task macro score by
$-0.05$ pp ($[-0.14,+0.04]$ pp) relative to Ordinary MoE.
Latest-write-only also remains within both frozen non-inferiority bounds.
Thus, both state-augmented variants satisfy the pre-specified SmolLM2
capability criterion.

On Gemma, the perplexity criterion is supported for both variants:
Persistent changes perplexity by $-0.06\%$
($[-0.45,+0.32]\%$), while latest-write-only changes it by $+0.30\%$
($[-0.18,+0.79]\%$).
The downstream macro contrasts are $-0.16$ pp
($[-1.66,+1.35]$ pp) for Persistent and $-0.49$ pp
($[-1.31,+0.33]$ pp) for latest-write-only.
Because both intervals cross the pre-specified $-1$ pp boundary, the
complete Gemma capability criterion is inconclusive rather than
supported.

These results establish a small architectural overhead and no detected
perplexity degradation beyond the frozen tolerance in either family,
while leaving downstream capability preservation on Gemma unresolved.

Matched single-GPU measurements show implementation-dependent wall-clock
effects with opposite directions across the two families; we therefore
report these measurements descriptively in Appendix~\ref{app:runtime}
rather than treating them as evidence of a general speedup or slowdown.

\subsection{Trained Routers Use the Persistent State}
\label{sec:results_state_use}

The routing ledger is only useful if trained routers actually rely on the
state pathway. We therefore measure direct router dependence with a local
same-state intervention. For a natural forward-pass pair
$(x_\ell,s_\ell)$, we attenuate only the state's contribution to the
router logits,
\begin{equation}
z_\ell(\lambda)
=
W_x^\ell x_\ell
+
\lambda W_s^\ell s_\ell
+
b^\ell,
\qquad
\lambda \in \{1,0.75,0.5,0.25,0\}.
\label{eq:state_suppression}
\end{equation}
The hidden representation and state are held fixed, and later model
computation is not rerun. This therefore measures the direct dependence
of the current routing decision on the explicit state rather than a
full-trajectory intervention.

\begin{figure*}[t]
\centering
\includegraphics[width=0.9\textwidth]{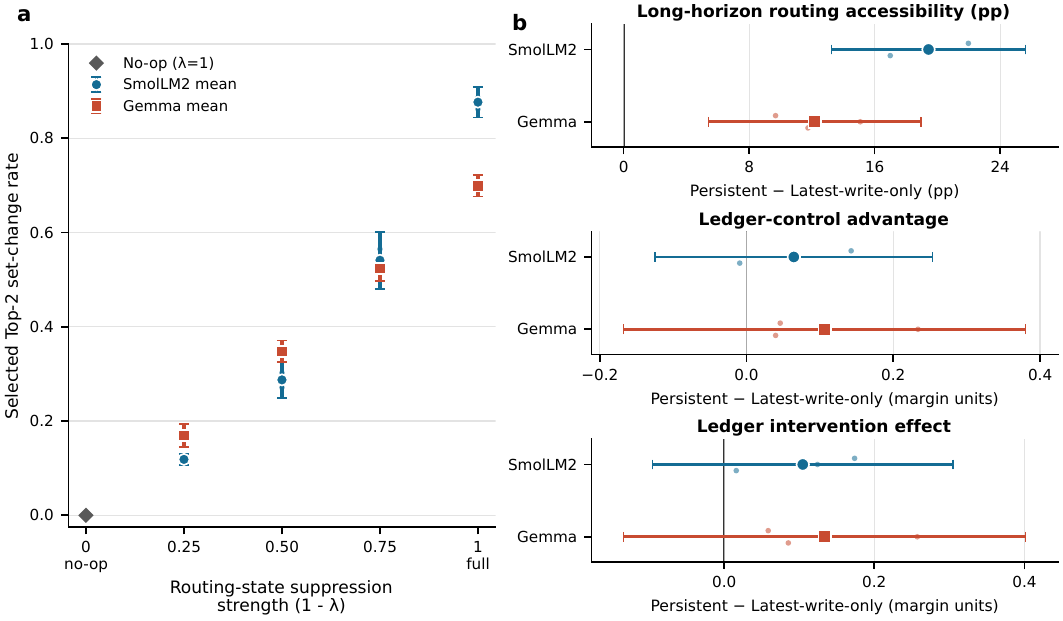}
\caption{
\textbf{State use and persistence across training seeds 42/43/44.}
\textbf{(a)} Persistent-only local state-suppression dose response from
Section~\ref{sec:results_state_use}.
\textbf{(b)} Persistent-minus-latest-write-only contrasts for the three
frozen persistence endpoints: long-horizon routing accessibility,
ledger-control advantage, and ledger intervention effect.
Points are three-seed means with Student-$t$ 95\% intervals
($n=3$, df$=2$); $\cdot$ marks individual seeds.
}
\label{fig:state-use-persistence}
\end{figure*}

Figure~\ref{fig:state-use-persistence}a shows a graded response as the
state contribution is removed. At full suppression
($\lambda=0$), the selected Top-2 expert set changes for a three-seed
mean of 87.6\% of SmolLM2 routing decisions and 69.9\% of Gemma
decisions. The corresponding seed-level intervals are
$[84.4,90.9]\%$ and $[67.6,72.2]\%$.
Thus, in both model families, the trained routers make substantial use
of the explicit routing state.

\subsection{Persistence Improves Long-Horizon Routing Accessibility}
\label{sec:results_persistence}

State use alone does not establish that cross-layer accumulation matters:
a router could rely on a state pathway whose useful information is effectively
local. We therefore compare Persistent SA-MoE with the matched
latest-write-only control.

Our primary persistence endpoint measures how well the native routing state
predicts routing four or eight MoE layers ahead. At the $10^6$-example probe
budget, we average transition-only source-cluster accuracy over the four
pre-specified horizon-4 and horizon-8 source--target pairs within each family.
Figure~\ref{fig:state-use-persistence}b reports the paired
Persistent-minus-latest-write-only contrasts across training seeds.

Persistent improves long-horizon routing accessibility by
$19.4$ pp on SmolLM2 (95\% CI $[13.2,25.6]$ pp) and
$12.2$ pp on Gemma ($[5.4,18.9]$ pp).
The contrast is positive in all three training seeds for both families.
Thus, persistence changes the information carried by the native routing state
in a way that remains visible many layers later.

The intervention-derived persistence contrasts in
Figure~\ref{fig:state-use-persistence}b are less uniform:
their three-seed means favor Persistent, but their confidence intervals cross
zero. We therefore treat long-horizon accessibility as the replicated evidence
for an effect of persistent accumulation, rather than claiming that persistence
improves every downstream intervention metric.

The retained history is also distributed across depth rather than dominated
by the most recent update. Across training seeds, non-recent writes account
for 91.7\% of absolute ledger contribution on SmolLM2 and 90.3\% on Gemma,
with effective source supports of 9.49 and 5.82 writes, respectively.
These quantities describe the distribution of absolute forward contribution;
they do not measure signed net support or downstream causal importance.
Full definitions and per-seed results are given in
Appendix~\ref{app:ledger_distribution}.

\begin{figure*}[t]
\centering
\includegraphics[width=\textwidth]{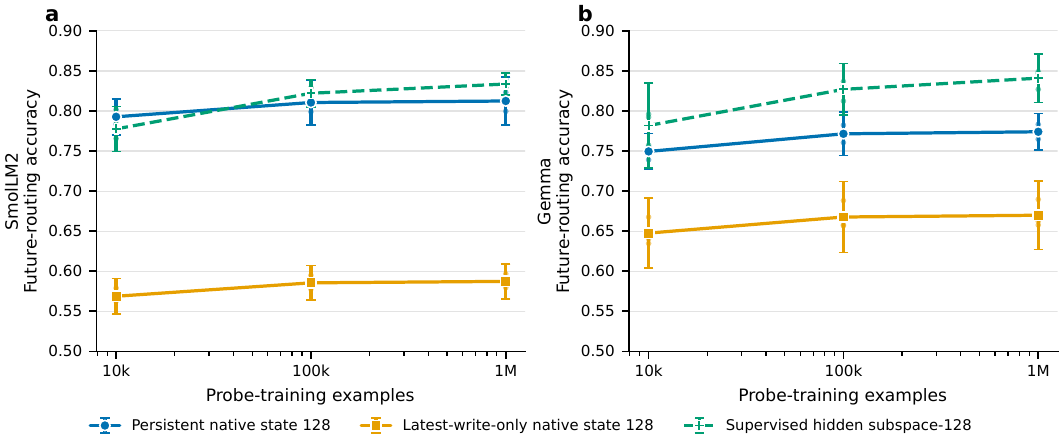}
\caption{
\textbf{Compact access to future routing across training seeds.}
Future-router Top-1 prediction accuracy versus probe-training budget for the
Persistent native state, matched latest-write-only native state, and a
supervised 128-dimensional subspace of the Persistent model's ordinary hidden
stream.
Points are three-seed means with Student-$t$ 95\% intervals
($n=3$, df$=2$); $\cdot$ marks individual seeds.
}
\label{fig:probe-curves}
\end{figure*}

\subsection{Compact Access to Future Routing}
\label{sec:results_probes}

We next ask whether the persistent state provides a compact representation of
future routing, rather than information that is uniquely available to SA-MoE.
Figure~\ref{fig:probe-curves} compares the native 128-dimensional routing
state with a supervised 128-dimensional representation learned from the same
Persistent model's ordinary hidden stream, across probe-training budgets from
$10^4$ to $10^6$ examples.

The Persistent native state substantially outperforms the matched
latest-write-only state at every probe budget in both model families.
At $10^6$ examples, the Persistent-minus-latest-write-only gap is
$22.5$ pp on SmolLM2 (95\% CI $[19.8,25.2]$ pp) and
$10.4$ pp on Gemma ($[5.3,15.5]$ pp).
This is consistent with the long-horizon persistence result in
Section~\ref{sec:results_persistence}, while using the full frozen probe
aggregation rather than only the pre-specified long-horizon endpoint.

The ordinary hidden stream nevertheless contains comparable routing
information. At the largest training budget, the supervised hidden
subspace reaches 83.4\% accuracy on SmolLM2 and 84.1\% on Gemma,
compared with 81.2\% and 77.4\% for the Persistent native state.
Thus, SA-MoE does not make future-routing information uniquely available.
Its architectural distinction is that a compact routing-specific
representation is already present in the model's native forward state,
without first learning a separate post-hoc representation.

\begin{figure*}[t]
\centering
\includegraphics[width=\textwidth]{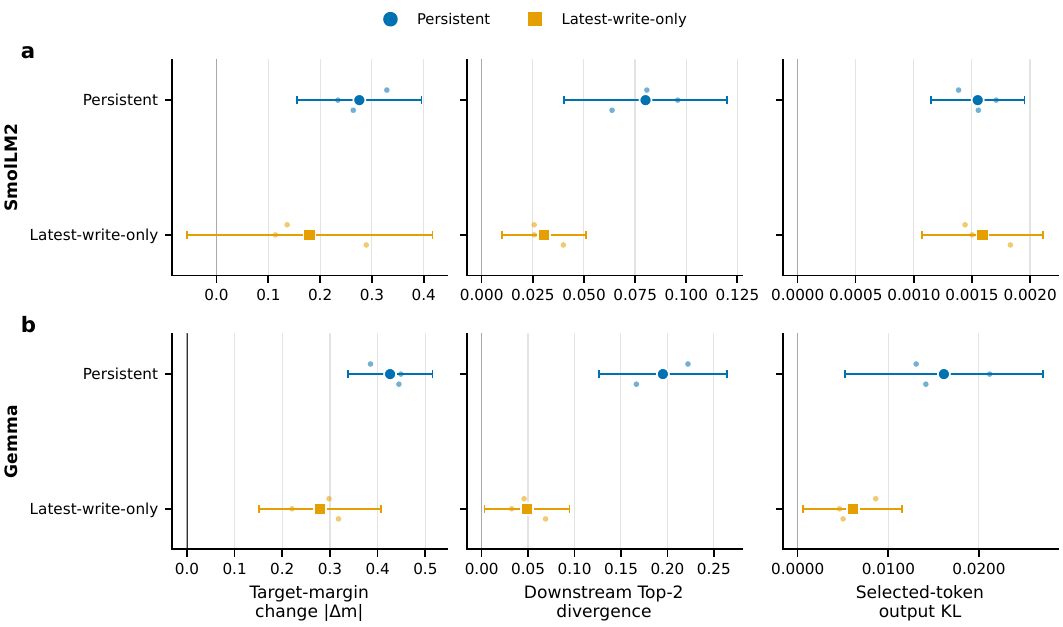}
\caption{
\textbf{Downstream consequences of ledger-selected source suppression.}
Effects on the audited Top-2 boundary, subsequent Top-2 routing, and the
selected-token output distribution for Persistent and latest-write-only.
Points are three-seed means with Student-$t$ 95\% intervals
($n=3$, df$=2$); $\cdot$ marks individual seeds.
}
\label{fig:ledger-downstream}
\end{figure*}

\subsection{Attribution Quality and Downstream Consequences}
\label{sec:results_attribution}

Exact forward provenance need not coincide with the source whose removal has
the largest downstream effect. We therefore compare clean-trajectory
attribution scores with exhaustive full-forward source interventions.
In the frozen seed-42 analysis, ledger scores correlate with intervention
effects more strongly than largest-write-norm: within-target Spearman
correlations are $0.369$ versus $0.187$ on SmolLM2 and $0.823$ versus
$0.341$ on Gemma. Gradient$\times$write is stronger still, reaching
$0.769$ and $0.890$, respectively. The same ordering holds for top-source
accuracy and normalized oracle regret (Appendix~\ref{app:attribution}).
This distinction is expected: the ledger records exact forward contribution
within the state pathway, whereas gradient$\times$write additionally
incorporates target-dependent downstream sensitivity.

Useful cross-layer attribution is also available without architectural
modification. On matched Ordinary MoEs, the \citet{li2026understanding} post-hoc attribution
achieves three-seed mean Spearman correlations of $0.570$ on SmolLM2 and
$0.620$ on Gemma; gradient$\times$component reaches $0.721$ and $0.782$.
These values are not directly comparable with the SA-MoE numbers because
state writes and Ordinary-MoE residual components are different audit
objects. Operationally, if the required forward-pass quantities are retained,
both ledger and Li-style scores can be materialized without a
target-dependent backward pass, while gradient-based scores require
target-dependent autograd. Phase-separated timing, memory, retention, and
reconstruction measurements are reported in
Appendix~\ref{app:attribution_costs}.

Finally, ledger-selected writes have consequences beyond the scalar margin
used to rank them (Figure~\ref{fig:ledger-downstream}).
Across training seeds, suppressing the Persistent ledger-selected write
changes the audited margin by $0.276$ on SmolLM2 and $0.426$ on Gemma,
produces downstream Top-2 divergences of $0.080$ and $0.195$, and yields
selected-token output KL divergences of $0.0016$ and $0.0162$,
respectively. Latest-write-only interventions also produce downstream
effects, and the Persistent-minus-latest-write-only contrasts are not
uniformly resolved across training seeds. These results show that
ledger-selected writes are computationally consequential, but do not imply
that exact provenance is an optimal causal ranking or that persistence
uniformly increases every intervention effect.

\section{Discussion and Limitations}
\label{sec:discussion}

The contribution of SA-MoE is not that a recorded recurrent state admits a
telescoping decomposition. For any recurrent trajectory $h_\ell$ and fixed
linear target readout $W_L$,
$W_L h_L = W_L h_0 + \sum_{\ell<L} W_L(h_{\ell+1}-h_\ell)$;
this observation also applies to nonlinear recurrent routers such as RMoE.
SA-MoE instead makes routing provenance an explicit interface of the forward
computation: a dedicated routing state is updated through a learned write
path, its realized post-clamp writes form the native audit object, and
individual writes have a direct source-level intervention semantics.
Empirically, trained routers use this pathway, persistence increases
long-horizon routing accessibility relative to replacement, and the resulting
history is distributed across earlier writes.

Exact provenance should also be distinguished from causal importance.
The ledger exactly decomposes the state-mediated component of a realized
routing margin, but it does not cover influences carried through the ordinary
hidden stream or guarantee the strongest intervention ranking.
Gradient$\times$write can rank consequential sources more strongly because it
incorporates downstream sensitivity, while \citet{li2026understanding}'s post-hoc method
recovers related cross-layer attribution in ordinary MoEs without architectural
modification. Likewise, sufficiently trained hidden-state representations can
match or exceed the native state's future-routing accuracy. The value of the
ledger is therefore its architecture-native routing-specific audit object,
not exclusive access to routing information or universal superiority over
post-hoc analysis.

Our latest-write-only control also does not hold state geometry fixed.
Its replacement state is bounded by the $\tanh$ write to $[-1,1]$, whereas
Persistent can accumulate toward the $[-5,5]$ clamp. The comparison therefore
tests accumulation versus replacement under the state distributions produced
by those mechanisms; differences in state magnitude and saturation cannot be
separated from accumulation itself. State-geometry diagnostics and additional
control details are reported in the appendix.

Finally, the experiments cover two sparsely upcycled model families and three
training seeds per architecture. This provides replication of the principal
effects but leaves wide uncertainty for some intervention contrasts and does
not establish behavior at larger MoE scales. More fundamentally, the ledger audits only the
persistent routing-state pathway: it does not assign semantic meaning to state
coordinates, provide a complete causal explanation of expert selection, or
imply improved generation quality or safety.

\section{Conclusion}

We introduced SA-MoE, an MoE architecture that makes one cross-layer routing
pathway explicitly auditable. A compact persistent state is updated by learned
write events, and the realized writes exactly decompose the state-mediated
contribution to later routing margins. Across two sparsely upcycled model
families and three training seeds, trained routers use this pathway,
persistence increases long-horizon routing accessibility relative to a matched
latest-write-only control, and the resulting ledger is distributed across
earlier writes. Full-forward suppression further shows that ledger-selected
writes can affect subsequent routing and model outputs.

The routing ledger is not a complete explanation of expert selection, nor is
exact provenance equivalent to optimal causal ranking. Post-hoc attribution
can recover related cross-layer influences without architectural modification,
and sensitivity-aware methods can identify stronger intervention targets.
SA-MoE instead provides a complementary design point: routing-specific
provenance is represented directly in the model computation as a compact state,
an explicit sequence of source events, and a native object for retrospective
query and intervention.

\subsection*{AI use statement}

We used generative AI tools to assist with experimental methodology,
interpretation and presentation of results, code generation and implementation,
and manuscript preparation. For the implementation work, generative AI was
used to help write, modify, and debug research code used in model training,
evaluation, probing, intervention analysis, and result processing. We also used
generative AI to draft and edit manuscript text, improve organization and
readability, check consistency between claims and reported results, identify
relevant literature, assist with literature searches and reference formatting,
and suggest titles, captions, and section structure.

All AI-assisted work was reviewed by the authors. AI-assisted code was inspected,
tested, and validated against expected behavior and experimental outputs before
being used for reported results. Numerical claims and interpretations were
checked against the underlying experimental results and figures, and suggested
citations and descriptions of prior work were verified against the original
sources before inclusion. Generative AI was not treated as a source of
experimental evidence, and the authors made the final decisions about the
methodology, implementation, analysis, claims, and manuscript content. We take
responsibility for the final content of this work, including text, code, claims,
and artifacts produced with the aid of generative AI.

\subsection*{Ethics statement}

This work studies model architecture and interpretability and does not involve
human participants, user studies, or the collection of new personal or
sensitive data. Our experiments use existing language models and research
datasets comprising web text, code, and mathematical content. These source
models and datasets may inherit biases, undesirable content, or other
limitations from their underlying corpora. SA-MoE is intended as a mechanism
for auditing one routing-specific computational pathway; the resulting
provenance should not be interpreted as a guarantee of complete
interpretability, model safety, or reliable behavior in deployment. We
therefore avoid making claims of improved safety or generation quality from
the intervention results reported in this work.

\subsection*{Reproducibility statement}

Section~\ref{sec:method} defines the SA-MoE architecture, realized routing
writes, and routing-ledger equations. Section~\ref{sec:setup}
specifies the model families, matched controls, evaluation roles,
intervention design, and statistical treatment used for the main results.
Appendix~\ref{app:training} provides dense-source revisions, MoE construction,
initialization, continued-pretraining settings, parameter and compute
accounting, runtime measurements, and artifact identities.
Appendix~\ref{app:evaluation} specifies validation and capability evaluation,
local state suppression, and the complete future-routing probe protocol.
Appendix~\ref{app:ledger_distribution} defines the Persistent ledger,
numerical reconstruction checks, distributed-history metrics, the distinct
latest-write-only audit object, and architecture-specific suppression
semantics. Appendix~\ref{app:attribution} gives deterministic target
selection, source-ranking methods, full-forward intervention metrics,
ordinary-MoE post-hoc attribution, and audit-cost measurements.
Finally, Appendix~\ref{app:seed_results} reports the individual results for
training seeds 42, 43, and 44 and explicitly identifies analyses that instead
use a fixed checkpoint or diagnostic pilot.
These specifications cover the procedures and statistical units underlying
the results in Sections~\ref{sec:results_quality}--%
\ref{sec:results_attribution}.

\bibliography{bibli}
\bibliographystyle{iclr2027_conference}

\appendix

\section{Model, Training, and Reproducibility}
\label{app:training}

This appendix gives the model construction, initialization, optimization,
analytical cost, and runtime details underlying the experiments in
Section~4. Evaluation and probe specifications are given in
Appendix~\ref{app:evaluation}; ledger-specific definitions and interventions
are given in Appendices~\ref{app:ledger_distribution}
and~\ref{app:attribution}.

\subsection{Dense Source Models and Tokenizer Provenance}
\label{app:dense_sources}

We sparse-upcycle two pretrained dense language models:
\texttt{HuggingFaceTB/SmolLM2-135M} and
\texttt{google/gemma-3-270m}. The local model and tokenizer bytes used by the
reported experiments are reproducible from the following public repository
revisions:
\begin{center}
\small
\begin{tabular}{lll}
\toprule
Family & Repository & Reproducible revision \\
\midrule
SmolLM2 &
\texttt{HuggingFaceTB/SmolLM2-135M} &
\texttt{93efa2f097d58c2a74874c7e644dbc9b0cee75a2} \\
Gemma &
\texttt{google/gemma-3-270m} &
\texttt{9b0cfec892e2bc2afd938c98eabe4e4a7b1e0ca1} \\
\bottomrule
\end{tabular}
\end{center}

For each family, the tokenizer is taken from the same repository revision as
the model; no separate tokenizer repository or tokenizer revision is used.
Verification against these revisions reproduces the frozen local model and
tokenizer bytes. We therefore describe them as byte-compatible reproducible
revisions, rather than as recovered records of the original download event.
The accompanying artifact manifests record the corresponding model,
tokenizer, configuration, and source-code SHA-256 hashes.

\subsection{MoE Architecture}
\label{app:model_configuration}

Every dense feed-forward block is replaced with an MoE block containing
16 routed experts and one always-active shared expert. All experts retain the
MLP geometry of the corresponding dense source layer. Each token executes the
shared expert and the Top-2 routed experts.

\begin{table}[t]
\centering
\small
\setlength{\tabcolsep}{5pt}
\caption{Primary model and MoE configuration. The routing-state entries apply
to Persistent and latest-write-only; Ordinary MoE has no routing state.}
\label{tab:appendix_model_configuration}
\begin{tabular}{lcc}
\toprule
 & SmolLM2 & Gemma \\
\midrule
Dense source & SmolLM2-135M & Gemma-3-270M \\
Transformer layers & 30 & 18 \\
Hidden size & 576 & 640 \\
Expert intermediate size & 1536 & 2048 \\
Routed experts / layer & 16 & 16 \\
Shared experts / layer & 1 & 1 \\
Routed experts / token & 2 & 2 \\
Active expert paths / token & 3 & 3 \\
Routing-state width $d_s$ & 128 & 128 \\
State clamp & $[-5,5]$ & $[-5,5]$ \\
Sequence length & 2048 & 2048 \\
Capacity factor & none & none \\
Token dropping & none & none \\
\bottomrule
\end{tabular}
\end{table}

For router logits $z_\ell$, the implementation first computes a softmax over
all 16 routed experts, selects the Top-2 entries, and renormalizes their
probabilities:
\begin{equation}
p_{\ell,i}
=
\operatorname{softmax}(z_\ell)_i,
\qquad
\widetilde p_{\ell,i}
=
\frac{p_{\ell,i}}
{\sum_{j\in \operatorname{Top2}(z_\ell)} p_{\ell,j}+10^{-8}}
\quad
(i\in\operatorname{Top2}(z_\ell)).
\end{equation}
The two selected routed-expert outputs are weighted by
$\widetilde p_{\ell,i}$ and summed; the always-active shared-expert output is
then added to this routed output. There is no capacity clipping or
dropped-token branch.

Ordinary MoE routes from $x_\ell$ only. Persistent and latest-write-only route
from the concatenation of $x_\ell$ and $s_\ell$. The implemented router is a
bias-free linear map; equivalently, the bias term in the general notation of
Section~3 is $b_\ell=0$ for the models reported here.

\subsection{Sparse Upcycling and Initialization}
\label{app:initialization}

Shared and routed experts are initialized as deep copies of the pretrained
dense MLP at the corresponding layer. We then partially reinitialize each
routed expert independently. For every routed-expert weight tensor with at
least two dimensions, each entry is selected independently with probability
$0.55$; selected entries are replaced by draws from a Gaussian whose mean and
standard deviation equal those of the original tensor. Routed-expert biases are
not reinitialized. The shared expert remains an unchanged copy of the
pretrained dense MLP.

Router weights use Kaiming-uniform initialization with parameter $a=0.01$.
Routers contain no bias. The state write head is initialized to produce an
exact zero write: both its weights and bias are initialized to zero. The
training seed is set before sparse upcycling and data construction.

The principal experiments use three independent training seeds,
$42$, $43$, and $44$, for every model family and architecture variant.
Architecture comparisons are paired by seed. The same construction algorithm
and dense source checkpoint are used for matched variants, but we do not claim
that historical initialized parameter tensors were retained byte-for-byte
across variants.

\subsection{Persistent and Latest-Write-Only State Implementations}
\label{app:state_implementation}

At the beginning of each model forward pass, the routing state is initialized
to zero in the hidden-stream dtype. It is then threaded through successive MoE
layers for each token. The write head reads the always-active shared-expert
representation $h^{\mathrm{shared}}_\ell$ and computes
\begin{equation}
u_\ell
=
\tanh\!\left(
W^\ell_w h^{\mathrm{shared}}_\ell + b^\ell_w
\right).
\end{equation}
The main text suppresses the write-head bias from the notation; the
implementation includes it and initializes it to zero.

For Persistent,
\begin{equation}
s_{\ell+1}
=
\operatorname{clip}
\left(
s_\ell+u_\ell,-5,5
\right).
\label{eq:appendix_persistent_state}
\end{equation}
For latest-write-only,
\begin{equation}
s_{\ell+1}
=
\operatorname{clip}
\left(
u_\ell,-5,5
\right).
\label{eq:appendix_lwo_state}
\end{equation}
There is no detached or separately retained accumulator in the
latest-write-only model. Because $u_\ell$ is produced by a $\tanh$, its native
replacement state lies in $[-1,1]$ even though the common outer clamp is
$[-5,5]$. The resulting state-geometry difference from Persistent is discussed
further in Appendix~\ref{app:state_geometry}.

The model state is stored in the forward dtype; the reported training runs use
bf16. Attribution calculations that require state differences or router-weight
dot products convert the relevant quantities to float32, as detailed in
Appendix~\ref{app:ledger_distribution}.

\subsection{Continued-Pretraining Recipe}
\label{app:training_recipe}

Training examples are sampled from FineWeb-Edu
\citep{penedo2024fineweb}, The Stack Dedup Python
\citep{thestack}, and FineMath-4+
\citep{benallal2025smollm2} with probabilities
$0.50$, $0.25$, and $0.25$, respectively. These are sampling probabilities
before packing and are not asserted to equal the realized proportions of the
packed token stream.

All reported runs use 30,000 optimizer steps and a global batch of
512 length-2048 sequences, corresponding to
$1{,}048{,}576$ packed input tokens per optimizer step and a nominal total of
\begin{equation}
30{,}000 \times 1{,}048{,}576
=
31{,}457{,}280{,}000
\end{equation}
input tokens per training run.

\begin{table}[t]
\centering
\small
\setlength{\tabcolsep}{5pt}
\caption{Continued-pretraining configuration. Microbatch and accumulation
differ where needed to fit the model family while preserving the same global
batch and token budget.}
\label{tab:appendix_training_configuration}
\begin{tabular}{lcc}
\toprule
 & SmolLM2 & Gemma \\
\midrule
Precision & bf16 & bf16 \\
Optimizer steps & 30,000 & 30,000 \\
Sequence length & 2048 & 2048 \\
World size & 16 & 16 \\
Global sequences / step & 512 & 512 \\
Tokens / optimizer step & 1,048,576 & 1,048,576 \\
Ordinary/Persistent microbatch / GPU & 8 & 2 \\
Ordinary/Persistent grad.\ accumulation & 4 & 16 \\
Latest-write-only microbatch / GPU & 8 & 4 \\
Latest-write-only grad.\ accumulation & 4 & 8 \\
Optimizer & AdamW & AdamW \\
$\beta_1,\beta_2$ & $0.9,0.95$ & $0.9,0.95$ \\
$\epsilon$ & $10^{-8}$ & $10^{-8}$ \\
Base peak learning rate $\eta$ & $3\times10^{-4}$ & $3\times10^{-4}$ \\
Router/write-head learning rate & $2\eta$ & $2\eta$ \\
Shared-expert learning rate & $0.1\eta$ & $0.1\eta$ \\
Shared-expert frozen period & 3,000 steps & 3,000 steps \\
Warmup & 1,500 steps & 1,500 steps \\
Post-warmup schedule & cosine to $0.1\times$ peak & cosine to $0.1\times$ peak \\
Weight decay & $0.1$ & $0.1$ \\
Global gradient-norm clip & $1.0$ & $1.0$ \\
Load-balance coefficient & $10^{-3}$ & $10^{-3}$ \\
Router z-loss coefficient & $10^{-3}$ & $10^{-3}$ \\
\bottomrule
\end{tabular}
\end{table}

Routed experts and remaining base-model parameters use the base learning rate.
The shared expert uses $0.1\eta$ and is held at zero learning rate for the
first 3,000 optimizer steps. Router and write-head parameters use $2\eta$.
After a 1,500-step linear warmup, each parameter group follows a cosine decay
to $0.1$ times its peak learning rate. Weight decay is applied to eligible
matrix weights but not to biases or normalization parameters.

Let $N=16$ denote the number of routed experts, $f_i$ the fraction of hard
Top-2 assignments to expert $i$, and $P_i$ its mean pre-Top-2 softmax
probability. The load-balancing term is
\begin{equation}
\mathcal{L}_{\mathrm{bal}}
=
10^{-3} N \sum_{i=1}^{N} f_i P_i ,
\end{equation}
where both selected Top-2 slots contribute to $f_i$. Each MoE layer also uses
the router z-loss~\citep{zoph2022}
\begin{equation}
\mathcal{L}_{z}
=
10^{-3}
\operatorname{mean}
\left[
\operatorname{logsumexp}(z_\ell)^2
\right].
\end{equation}
Auxiliary losses are averaged across MoE layers and added to the language-model
cross-entropy objective.

The Gemma latest-write-only runs use microbatch 4 with gradient accumulation
8 rather than the microbatch-2, accumulation-16 shape used by the
Ordinary/Persistent Gemma runs. This changes the per-device execution shape
but preserves the world size, global batch, sequences per optimizer step, and
nominal token budget.

\subsection{Parameter Counts and Analytical Compute}
\label{app:architecture_cost}

Table~\ref{tab:appendix_architecture_cost} reports the analytical model-size
and active-compute quantities used in Section~5.1. Active parameters count all
non-routed parameters, the always-active shared expert, and the two selected
routed experts. Forward FLOPs count the reported analytical matrix-multiplication
work per token and should not be interpreted as measured wall-clock latency.

\begin{table}[t]
\centering
\small
\setlength{\tabcolsep}{4.5pt}
\caption{Model size and analytical active compute. Parameter counts include
each unique parameter once. Values are shown at higher precision than in the
main-text table.}
\label{tab:appendix_architecture_cost}
\begin{tabular}{llrrr}
\toprule
Family & Architecture & Params. & Active/token & Fwd.\ FLOPs/token \\
\midrule
SmolLM2 & Ordinary
& 1.4088B & 294.044M & 729.575M \\
& Latest-write-only
& 1.4111B & 296.321M & 734.122M \\
& Persistent
& 1.4111B & 296.321M & 734.122M \\
\midrule
Gemma & Ordinary
& 1.4007B & 409.840M & 876.192M \\
& Latest-write-only
& 1.4023B & 411.354M & 879.215M \\
& Persistent
& 1.4023B & 411.354M & 879.215M \\
\bottomrule
\end{tabular}
\end{table}

Relative to Ordinary MoE, Persistent therefore increases the analytical
forward FLOP count by approximately $0.62\%$ for SmolLM2 and $0.34\%$ for
Gemma. Persistent and latest-write-only have identical parameter and
analytical active-compute counts within a family.

\subsection{Common-Hardware Runtime Measurements}
\label{app:runtime}

We additionally measured full-sequence runtime on a common NVIDIA GH200 120GB
GPU. These measurements are descriptive implementation diagnostics rather than
training-seed replications. Inference measurements are full length-2048 forward
passes rather than autoregressive cached decoding. The training measurement is
a single-GPU forward--backward pass and excludes optimizer updates and
distributed communication.

\begin{table}[t]
\centering
\small
\setlength{\tabcolsep}{5pt}
\caption{Common-hardware median runtime in milliseconds on one NVIDIA GH200
120GB. B1/B16 denote inference batch size; Train B4 is a forward--backward
training step with batch size 4. These timings are descriptive and are not used
as statistical architecture claims.}
\label{tab:appendix_runtime}
\begin{tabular}{llrrr}
\toprule
Family & Architecture & Inf.\ B1 & Inf.\ B16 & Train B4 \\
\midrule
SmolLM2 & Ordinary & 117.3 & 192.2 & 446.8 \\
& Latest-write-only & 151.4 & 210.7 & 545.8 \\
& Persistent & 140.9 & 204.2 & 524.4 \\
\midrule
Gemma & Ordinary & 94.3 & 191.8 & 318.4 \\
& Latest-write-only & 82.2 & 176.9 & 326.6 \\
& Persistent & 82.5 & 182.3 & 316.4 \\
\bottomrule
\end{tabular}
\end{table}

The direction of the measured runtime difference is not consistent across the
two model families. We therefore use the analytical FLOP counts as the
architecture-level compute accounting and report the wall-clock measurements
only as implementation-specific descriptive results.

\subsection{Source Freezes and Artifact Identity}
\label{app:reproducibility_artifacts}

The canonical Persistent and Ordinary-MoE implementation freeze is tied to
Git commit
\texttt{984f88b7f1aac9373a55d221154308cf82e1a2b8},
with an empty tracked-diff hash in the corresponding freeze manifest.
The latest-write-only runs are accompanied by launch manifests that bind the
model source, baseline source, configuration, upcycling code, trainer,
software environment, and run configuration to cryptographic hashes.

Result and configuration artifacts use SHA-256 identities over their exact
serialized bytes. The training-seed registry additionally records a structured
content hash computed from compact canonical JSON with sorted keys, comma/colon
separators, UTF-8 encoding, and no trailing newline; the human-readable
on-disk registry is stored as pretty, sorted JSON with a trailing newline.
These identities separate the logical checkpoint record from filenames or
storage locations.

The main replicated architecture-level results use training seeds
$42$, $43$, and $44$, paired by seed within each model family. Analyses that
use a fixed checkpoint rather than all three training realizations are labeled
explicitly at the point of use in the later appendices.

\section{Evaluation Protocols and Future-Routing Probes}
\label{app:evaluation}

This appendix specifies the evaluation pools, capability metrics, local
state-suppression analysis, and future-routing probes used in Sections~4
and~5. Training and model-construction details are given in
Appendix~\ref{app:training}; ledger and intervention definitions are given in
Appendices~\ref{app:ledger_distribution} and~\ref{app:attribution}.

\subsection{Analysis Roles and Evaluation Pools}
\label{app:evaluation_roles}

We separate analysis construction from final evaluation. Construction data are
used to choose layer sets, representations, controls, and aggregation rules;
where model selection is required, a separate calibration role is used.
Corresponding final-evaluation examples are not inspected until those choices
are frozen. Throughout the paper, ``held out'' refers to this analysis-role
separation. It does not imply that the underlying documents were absent from
the pretraining corpus of the original dense source model.

The language-model and local routing-state analyses use a balanced
three-domain final-evaluation pool. For each model family, it contains
96 sequences of length 2048:
32 prose sequences from the local FineWeb-Edu sample,
32 code sequences from The Stack Dedup Python, and
32 mathematics sequences from FineMath-4+.
Stable sequence identities are retained across matched architecture variants
so that paired comparisons operate on the same source material.

The full-forward attribution experiment uses a smaller, independently fixed
subset of this evaluation material. Its sequence and target-selection
procedure is specified in Appendix~\ref{app:attribution}.

\subsection{Validation Perplexity}
\label{app:perplexity}

The validation-perplexity quantity reported in Section~5.1 is computed on the
96-sequence balanced-domain pool described above; it is not perplexity on a
standard external validation corpus.

Packed examples contain pre-aligned next-token labels. No additional label
shift is applied during evaluation. Each length-2048 packed sequence contains
one aligned label for each of its 2048 input positions, including the final
input position, so each domain contributes exactly
\[
32 \times 2048 = 65{,}536
\]
valid labels and the full evaluation contains
\[
3 \times 65{,}536 = 196{,}608
\]
valid labels per model evaluation.

Let $\mathcal{T}$ denote the set of valid token positions and
$\ell_t$ the next-token negative log likelihood at position $t$.
We compute
\begin{equation}
\operatorname{NLL}
=
\frac{1}{|\mathcal{T}|}
\sum_{t\in\mathcal{T}} \ell_t,
\qquad
\operatorname{PPL}
=
\exp(\operatorname{NLL}).
\end{equation}
Thus the point estimate is token weighted, both within and across domains.
We do not compare absolute perplexities between SmolLM2 and Gemma because the
two model families use different tokenizers.

For each architecture contrast and training seed, the same 96 source
sequences are evaluated under both matched variants. Relative perplexity is
formed as
\begin{equation}
r_s
=
\frac{\operatorname{PPL}_{\mathrm{variant},s}}
     {\operatorname{PPL}_{\mathrm{Ordinary},s}},
\end{equation}
where $s\in\{42,43,44\}$ is the training seed. Cross-seed inference is
performed on $\log r_s$; the mean and Student-$t$ interval are transformed
back to the ratio scale. Percentage changes reported in the main text are
$100(r-1)$. Training seed, rather than token count or bootstrap replicate, is
the replication unit for these architecture-level claims.

\subsection{Downstream Capability Evaluation}
\label{app:capability}

We evaluate HellaSwag, ARC-Easy, PIQA, and WinoGrande zero-shot
(\texttt{num\_fewshot=0}), without a system instruction, chat template, or
free-form generation. Request construction follows the corresponding
EleutherAI \texttt{lm-evaluation-harness} task definitions at Git commit
\texttt{b954108c9baaaa934b4ad842033b31a97ee30816}. Our local request
construction and per-example metric implementation were checked against these
pinned task definitions before the canonical evaluations.

For HellaSwag, ARC-Easy, and PIQA, each candidate answer is scored by its
conditional continuation log likelihood. Raw accuracy selects the candidate
with largest log likelihood. For normalized accuracy, the conditional log
likelihood is divided by the Python character length of the corresponding
task choice before selection. Ties are resolved by the lowest candidate index.

WinoGrande uses its task-specific harness construction rather than treating
the candidate word itself as an independently scored continuation. Each
candidate is inserted into the task prefix and the conditional likelihood of
the remaining suffix is scored; the higher-scoring completed construction is
selected.

The primary task metrics are
\begin{itemize}
    \item normalized accuracy for HellaSwag;
    \item normalized accuracy for ARC-Easy;
    \item normalized accuracy for PIQA; and
    \item raw accuracy for WinoGrande.
\end{itemize}
The reported macro score is their unweighted arithmetic mean:
\begin{equation}
\operatorname{Macro}
=
\frac{1}{4}
\left(
\operatorname{AccNorm}_{\mathrm{HellaSwag}}
+
\operatorname{AccNorm}_{\mathrm{ARC\text{-}Easy}}
+
\operatorname{AccNorm}_{\mathrm{PIQA}}
+
\operatorname{Acc}_{\mathrm{WinoGrande}}
\right).
\end{equation}

The evaluated splits and complete example counts are shown in
Table~\ref{tab:capability_protocol}. Inputs are limited to 2048 tokens.
When truncation is required, context is left-truncated while retaining the
complete scored continuation and at least one context token.

\begin{table}[t]
\centering
\small
\setlength{\tabcolsep}{6pt}
\caption{Downstream capability evaluation protocol.}
\label{tab:capability_protocol}
\begin{tabular}{lrrl}
\toprule
Task & Split & Examples & Primary metric \\
\midrule
HellaSwag & validation & 10,042 & normalized accuracy \\
ARC-Easy & test & 2,376 & normalized accuracy \\
PIQA & validation & 1,838 & normalized accuracy \\
WinoGrande & \texttt{winogrande\_xl} validation & 1,267 & raw accuracy \\
\bottomrule
\end{tabular}
\end{table}

Table~\ref{tab:capability_task_contrasts} gives the three-seed mean
task-level changes relative to the matched Ordinary MoE. Values are absolute
percentage-point changes in the corresponding primary metric. Full
seed-by-seed values and Student-$t$ intervals are given in
Appendix~\ref{app:seed_results}.

\begin{table}[t]
\centering
\small
\setlength{\tabcolsep}{4.5pt}
\caption{Three-seed mean task-level capability changes relative to matched
Ordinary MoE, in percentage points. Macro is the unweighted four-task mean.
These are paired architecture contrasts, not differences between model
families.}
\label{tab:capability_task_contrasts}
\begin{tabular}{llrrrrr}
\toprule
Family & Architecture &
HellaSwag & ARC-Easy & PIQA & WinoGrande & Macro \\
\midrule
SmolLM2
& Latest-write-only & $+0.14$ & $-0.36$ & $+0.29$ & $-0.16$ & $-0.02$ \\
& Persistent        & $+0.01$ & $-0.51$ & $+0.36$ & $-0.08$ & $-0.05$ \\
\midrule
Gemma
& Latest-write-only & $-0.39$ & $-1.09$ & $-0.33$ & $-0.16$ & $-0.49$ \\
& Persistent        & $-0.44$ & $-0.36$ & $+0.07$ & $+0.11$ & $-0.16$ \\
\bottomrule
\end{tabular}
\end{table}

\subsection{Local Routing-State Suppression}
\label{app:state_suppression}

The state-use experiment in Section~\ref{sec:results_state_use} is a local
same-state counterfactual rather than a full-forward intervention. For each
natural Persistent forward pass, we retain the hidden representation
$x_\ell$ and routing state $s_\ell$ entering a router and recompute only that
router under
\begin{equation}
z_\ell(\lambda)
=
W_x^\ell x_\ell
+
\lambda W_s^\ell s_\ell
+
b^\ell,
\qquad
\lambda\in\{1,0.75,0.5,0.25,0\}.
\end{equation}
As noted in Appendix~\ref{app:model_configuration}, the implemented routers
are bias-free, so $b^\ell=0$ in the reported models.

Neither $x_\ell$ nor $s_\ell$ is recomputed after attenuation, and the
modified expert selection is not propagated through later model layers. The
primary metric is
\begin{equation}
\operatorname{Top2Change}(\lambda)
=
\Pr\!\left[
\operatorname{Top2}(z_\ell(\lambda))
\neq
\operatorname{Top2}(z_\ell(1))
\right],
\end{equation}
where Top-2 is treated as an unordered selected-expert set. The no-op endpoint
$\lambda=1$ is zero by definition, and $\lambda=0$ removes the direct state
term from the current router while leaving the naturally produced hidden
representation and state fixed.

The analysis uses the 96-sequence final-evaluation pool from
Appendix~\ref{app:evaluation_roles}. We first reduce the routing decisions
within a checkpoint to one checkpoint-level rate. Cross-checkpoint summaries
then treat training seeds 42, 43, and 44 as the three replicates and report
Student-$t$ 95\% intervals with two degrees of freedom. The full-suppression
Persistent seed-level rates are
\[
(0.8816,\;0.8861,\;0.8617)
\]
for SmolLM2 and
\[
(0.6912,\;0.7093,\;0.6967)
\]
for Gemma, giving the main-text means of $87.6\%$ and $69.9\%$,
respectively.

\subsection{Independent Future-Routing Probes}
\label{app:probes}

For each frozen source--target layer pair, a probe predicts the clean Top-1
expert index of the target router from a representation captured at the same
token position at an earlier source layer. The source representation is the
actual representation available at the source router: the general hidden
stream for hidden-state probes and the 128-dimensional native routing state
for native-state probes.

Training, calibration, and final-evaluation roles are disjoint. The frozen
probe-training budgets are
\[
10^4,\qquad 10^5,\qquad 10^6
\]
association-role examples. Calibration and final evaluation each contain
98,304 examples per source layer, balanced as 32,768 examples from each of
the prose, code, and mathematics domains. Each domain contribution comprises
64 selected token positions from each of 512 sequences. The final-evaluation
role is not loaded until representation construction, hyperparameter
selection, and fitted-model identities are frozen.

The frozen source and target layers are shown in
Table~\ref{tab:probe_pairs}.

\begin{table}[t]
\centering
\small
\setlength{\tabcolsep}{8pt}
\caption{Frozen source--target layer pairs for future-routing prediction.
Horizons count MoE-layer distance. Expert indices are local to each target
router; equal numeric indices at different layers do not denote the same
physical expert.}
\label{tab:probe_pairs}
\begin{tabular}{lccc}
\toprule
Family / source & Horizon 1 & Horizon 4 & Horizon 8 \\
\midrule
SmolLM2, source 7  & $7\!\rightarrow\!8$  & $7\!\rightarrow\!11$ & $7\!\rightarrow\!15$ \\
SmolLM2, source 14 & $14\!\rightarrow\!15$ & $14\!\rightarrow\!18$ & $14\!\rightarrow\!22$ \\
Gemma, source 4    & $4\!\rightarrow\!5$  & $4\!\rightarrow\!8$  & $4\!\rightarrow\!12$ \\
Gemma, source 8    & $8\!\rightarrow\!9$  & $8\!\rightarrow\!12$ & $8\!\rightarrow\!16$ \\
\bottomrule
\end{tabular}
\end{table}

\subsection{Probe Representations and Fit Selection}
\label{app:probe_fitting}

We evaluate the native 128-dimensional routing state and five representations
of the model's ordinary hidden stream:
\begin{enumerate}
    \item a full-hidden multinomial linear predictor;
    \item a one-hidden-layer MLP with a 128-unit ReLU bottleneck;
    \item a supervised 128-dimensional hidden subspace;
    \item a 128-dimensional PCA representation; and
    \item a 128-dimensional random projection, averaged over five
    independently fixed projections.
\end{enumerate}

Unless explicitly stated otherwise, a hidden-state representation is taken
from the same architectural variant being evaluated. In particular, the
``Persistent hidden subspace-128'' curve compared with the Persistent native
state in Figure~\ref{fig:probe-curves} is learned from the
\emph{Persistent model's ordinary hidden stream}; it is not an Ordinary-MoE
baseline.

Captured hidden and native-state features are stored from the natural forward
pass and converted to float32 before fitting. A \texttt{StandardScaler} is
fit on probe-training examples only and then applied to calibration and final
examples. Learned projected representations receive a second scaler fit only
on their projected training features.

For the linear predictors, the candidate regularization values are
\[
C\in\{0.01,0.1,1.0,10.0\}.
\]
Fitting uses scikit-learn 1.9.0 multinomial logistic regression with
\texttt{solver=lbfgs}, \texttt{max\_iter=1000}, and
\texttt{tol=1e-4}. Calibration macro-F1 selects the candidate; ties are
resolved in favor of smaller $C$.

For the nonlinear bottleneck, the candidate pairs are
\[
(\alpha,\eta)
\in
\{
(10^{-3},3\times10^{-4}),
(10^{-3},10^{-3}),
(10^{-4},3\times10^{-4}),
(10^{-4},10^{-3})
\}.
\]
The classifier has one 128-unit ReLU hidden layer and uses Adam with batch
size 256, \texttt{max\_iter=1000}, \texttt{tol=1e-4},
\texttt{n\_iter\_no\_change=20}, \texttt{early\_stopping=False}, and
deterministic shuffling. Calibration macro-F1 again selects the candidate.
The frozen candidate order implements the tie rule of larger $\alpha$ first
and then smaller learning rate.

PCA uses randomized SVD with 128 components and a deterministic seed. The
supervised hidden subspace is constructed from the row space of the fitted
full-hidden linear classifier: the classifier coefficient matrix is converted
to float64, decomposed by SVD, and deterministically orthogonally completed to
128 dimensions where necessary. The resulting 128-dimensional representation
is then used as the input to the final probe. Random-projection results average
five independently fixed 128-dimensional projections.

All frozen source-layer fits used for the reported analyses converged before
final evaluation. Hyperparameter selection uses calibration data only; final
evaluation is not used to choose a representation or fitting candidate.

\subsection{Transition-Only Diagnostic}
\label{app:transition_only}

Figure~\ref{fig:probe-curves} reports a diagnostic subset called
\emph{transition-only}. This term refers only to a transition in numeric
expert-index sets. Let
\[
E_s = \{e^{(1)}_s,e^{(2)}_s\},
\qquad
E_t = \{e^{(1)}_t,e^{(2)}_t\}
\]
be the clean Top-2 numeric expert indices at source and target layers for one
example. After sorting each pair, the example is retained exactly when
\begin{equation}
\operatorname{sort}(E_s)
\neq
\operatorname{sort}(E_t).
\label{eq:transition_only}
\end{equation}
Top-2 order is therefore ignored.

Because experts at different layers are distinct modules, equality of a
numeric index across layers does not imply persistence of a physical or
semantic expert identity. Conversely, the mask in
Equation~\ref{eq:transition_only} should not be interpreted as a semantic
expert-change label. It is a fixed numeric-index diagnostic intended to remove
examples on which the source and target Top-2 index sets happen to be
identical.

For retained examples, transition-only token accuracy is ordinary Top-1
classification accuracy:
\begin{equation}
A_{\mathrm{trans}}
=
\frac{
\sum_i
\mathbf{1}[i\in\mathcal{M}]
\mathbf{1}[\hat e_i=e_i]
}{
|\mathcal{M}|
},
\end{equation}
where $\mathcal{M}$ is the mask defined above, $e_i$ is the clean target
Top-1 numeric expert index, and $\hat e_i$ is the probe prediction.
The metric is undefined only if $\mathcal{M}$ is empty; no reported frozen
pair has an empty final-evaluation mask.

\subsection{Probe Aggregation and the Two Persistence Summaries}
\label{app:probe_aggregation}

Two related but intentionally different probe aggregates appear in the main
paper. We define them separately here to avoid conflating their numerical
values.

\paragraph{Long-horizon persistence endpoint.}
The pre-specified persistence endpoint in
Section~\ref{sec:results_persistence} and
Figure~\ref{fig:state-use-persistence}b uses only horizons 4 and 8 at the
$10^6$-example probe-training budget. Thus, for each family,
\[
\mathcal{P}_{\mathrm{long}}
=
\{
(s_1,t_{1,4}),
(s_1,t_{1,8}),
(s_2,t_{2,4}),
(s_2,t_{2,8})
\},
\]
containing four source--target pairs.

For training seed $r$, let
$A^{P}_{p,r}$ and $A^{L}_{p,r}$ denote the final transition-only native-state
accuracy for Persistent and latest-write-only on pair $p$. The seed-level
endpoint is
\begin{equation}
\Delta_{\mathrm{long},r}
=
\frac{1}{4}
\sum_{p\in\mathcal{P}_{\mathrm{long}}}
\left(
A^{P}_{p,r}
-
A^{L}_{p,r}
\right).
\label{eq:long_horizon_access}
\end{equation}
Training seeds are then treated as the replication unit.

The resulting seed-level contrasts, in percentage points, are
\begin{center}
\small
\begin{tabular}{lrrrr}
\toprule
Family & Seed 42 & Seed 43 & Seed 44 & Mean [95\% CI] \\
\midrule
SmolLM2 &
21.97 & 19.33 & 16.99 &
19.43 $[13.24,25.62]$ \\
Gemma &
9.68 & 15.09 & 11.75 &
12.17 $[5.40,18.94]$ \\
\bottomrule
\end{tabular}
\end{center}
These are the persistence values reported in
Section~\ref{sec:results_persistence}. In particular, the previously reported
21.97 pp and 9.68 pp values are the seed-42 entries, not the three-seed means.

\paragraph{Figure~\ref{fig:probe-curves} learning-curve aggregate.}
The probe-learning curves instead use the equal-pair mean over
\emph{all six} frozen source--target pairs in
Table~\ref{tab:probe_pairs}, including horizons 1, 4, and 8. For
representation $q$, budget $B$, and training seed $r$,
\begin{equation}
A_{q,B,r}
=
\frac{1}{6}
\sum_{p\in\mathcal{P}_{\mathrm{all}}}
A_{q,B,r,p}.
\label{eq:probe_curve_aggregate}
\end{equation}
Any frozen probe-fit reduction is performed within checkpoint before this
equal-pair mean; the three training checkpoints remain the replication units
for the Student-$t$ intervals shown in the figure.

At the $10^6$-example budget, the corresponding Persistent-minus-LWO
native-state contrasts are
\[
22.5\ \text{pp}
\quad\text{for SmolLM2},
\qquad
10.4\ \text{pp}
\quad\text{for Gemma}.
\]
These differ from Equation~\ref{eq:long_horizon_access} because the
Figure~\ref{fig:probe-curves} aggregate additionally includes the two
horizon-1 pairs.

Table~\ref{tab:probe_learning_curve_values} gives the numerical three-seed
means underlying the principal curves in Figure~\ref{fig:probe-curves}.

\begin{table}[t]
\centering
\small
\setlength{\tabcolsep}{5pt}
\caption{Three-seed mean transition-only future-routing accuracy (\%) for the
principal probe representations. ``P hidden-128'' is the supervised
128-dimensional subspace learned from the Persistent model's ordinary hidden
stream. Values are equal-pair means over all six frozen source--target pairs.}
\label{tab:probe_learning_curve_values}
\begin{tabular}{llrrr}
\toprule
Family & Representation & 10k & 100k & 1M \\
\midrule
SmolLM2
& Persistent native state & 79.28 & 81.06 & 81.25 \\
& Latest-write-only native state & 56.87 & 58.55 & 58.73 \\
& P hidden-128 & 77.76 & 82.21 & 83.37 \\
\midrule
Gemma
& Persistent native state & 74.95 & 77.15 & 77.41 \\
& Latest-write-only native state & 64.76 & 66.77 & 66.99 \\
& P hidden-128 & 78.19 & 82.70 & 84.10 \\
\bottomrule
\end{tabular}
\end{table}

Thus the native Persistent state retains a large advantage over the matched
latest-write-only state at each probe budget, while the supervised
representation of the Persistent model's ordinary hidden stream can match or
exceed the native state with sufficient probe training. The latter comparison
is why we interpret the native state as providing compact, directly available
routing information rather than information that is unique to the
routing-state pathway.

\subsection{Additional Representation Controls}
\label{app:additional_probe_controls}

The full representation study additionally evaluates the full hidden-state
linear predictor, 128-unit nonlinear bottleneck, PCA-128, and five-replicate
random-projection-128 controls defined in
Appendix~\ref{app:probe_fitting}. These controls address different possible
explanations for the performance of the supervised hidden subspace:
unrestricted linear accessibility, nonlinear probe capacity, unsupervised
low-dimensional variance, and arbitrary low-dimensional projection.

The representation comparison is not parameter matched to the native routing
state: a learned predictor applied to a hidden representation may contain more
fitted parameters than the model's native 128-dimensional state pathway.
Accordingly, we use these probes to characterize accessibility rather than to
claim an equal-resource competition between representations. The complete
per-seed numerical probe record is reported in
Appendix~\ref{app:seed_results}.

\section{Ledger Construction and Persistence Diagnostics}
\label{app:ledger_distribution}

This appendix specifies the native Persistent routing ledger, its numerical
reconstruction checks, the distributional summaries used in
Section~\ref{sec:results_persistence}, and the distinct direct-state audit
object used for latest-write-only. Full-forward target selection, attribution
rankings, intervention outcomes, and audit-cost comparisons are given in
Appendix~\ref{app:attribution}.

\subsection{Native Persistent Ledger and Numerical Reconstruction}
\label{app:native_ledger}

The multi-write routing ledger is defined only for Persistent. For consecutive
natural states entering successive routers, the realized source write is
\begin{equation}
\Delta s_\ell
=
s_{\ell+1}-s_\ell .
\label{eq:appendix_realized_write}
\end{equation}
This is the state change that remains after the source layer's
$\tanh$ write proposal, accumulation with the incoming state, and elementwise
clamp. We never substitute the pre-clamp proposal $u_\ell$ for
$\Delta s_\ell$.

Consider a target layer $t$, with clean Top-1 expert $e_1$, clean
second-selected expert $e_2$, and clean best-excluded expert $e_3$.
Let $W^t_S$ denote the columns of the target router acting on the
128-dimensional state. We retain two margin decompositions:
\begin{align}
c^{\mathrm{top1}}_{\ell,t}
&=
\Delta s_\ell^\top
\left(
W^t_S[e_1]-W^t_S[e_2]
\right),
\\
c^{\mathrm{set}}_{\ell,t}
&=
\Delta s_\ell^\top
\left(
W^t_S[e_2]-W^t_S[e_3]
\right).
\label{eq:appendix_set_contribution}
\end{align}
The second quantity is the contribution to the clean Top-2 membership
boundary used throughout the attribution analyses.

Because $s_0=0$,
\begin{align}
\sum_{\ell<t}c^{\mathrm{top1}}_{\ell,t}
&=
s_t^\top
\left(
W^t_S[e_1]-W^t_S[e_2]
\right),
\\
\sum_{\ell<t}c^{\mathrm{set}}_{\ell,t}
&=
s_t^\top
\left(
W^t_S[e_2]-W^t_S[e_3]
\right).
\label{eq:appendix_ledger_reconstruction}
\end{align}
These equalities concern the direct state-mediated component of the target
router logits. Influences transmitted through the ordinary hidden stream are
outside this decomposition.

The model forward pass uses bf16. Natural states and router logits are
captured at the actual router hook and converted to float32 for attribution;
the relevant router weights are likewise copied to float32. The attribution
cache stores float32 realized-write vectors, target logits, and contribution
tensors. Reconstruction is therefore exact algebraically but need not be
bit-exact after finite-precision capture, subtraction, and summation.

Table~\ref{tab:ledger_reconstruction} reports the frozen numerical validation
on the construction checkpoints. The acceptance tolerance was
$10^{-4}$ absolute error. Both families remain more than an order of
magnitude below this threshold.

\begin{table}[t]
\centering
\small
\setlength{\tabcolsep}{7pt}
\caption{Numerical reconstruction of the Persistent ledger on the frozen
construction checkpoints. Errors are maximum absolute residuals between the
summed write contributions and the corresponding direct target-state margin
term. This is a numerical validation of the descriptive decomposition, not an
intervention result.}
\label{tab:ledger_reconstruction}
\begin{tabular}{lrr}
\toprule
Family & Top-1 margin & Top-2 boundary \\
\midrule
SmolLM2 & $3.815\times10^{-6}$ & $3.785\times10^{-6}$ \\
Gemma   & $2.623\times10^{-6}$ & $2.861\times10^{-6}$ \\
\bottomrule
\end{tabular}
\end{table}

\subsection{Distributed-History Metrics}
\label{app:history_metrics}

The distributed-history analysis uses the selected-set contribution
$c^{\mathrm{set}}_{\ell,t}$ from
Equation~\ref{eq:appendix_set_contribution}. For brevity, write
$c_{\ell,t}=c^{\mathrm{set}}_{\ell,t}$ and define the absolute attribution
mass for target $t$ as
\begin{equation}
A_t
=
\sum_{\ell<t}|c_{\ell,t}|.
\end{equation}
For $A_t>0$, define normalized absolute source weights
\begin{equation}
p_{\ell,t}
=
\frac{|c_{\ell,t}|}{A_t},
\qquad
\sum_{\ell<t}p_{\ell,t}=1.
\label{eq:absolute_source_weights}
\end{equation}

Let $r(t)$ denote the most recent eligible source before target $t$.
The \emph{non-recent absolute contribution share} is
\begin{equation}
R_t
=
\sum_{\substack{\ell<t\\\ell\neq r(t)}}p_{\ell,t}
=
1-p_{r(t),t}.
\label{eq:nonrecent_share}
\end{equation}
Thus ``non-recent'' means every eligible source except the immediately
preceding source; it does not denote a depth tercile or a fixed number of
layers.

We quantify concentration with inverse Herfindahl concentration:
\begin{equation}
N_{\mathrm{eff},t}
=
\frac{1}{\sum_{\ell<t}p_{\ell,t}^{\,2}}
=
\frac{A_t^2}
{\sum_{\ell<t}|c_{\ell,t}|^2}.
\label{eq:effective_support}
\end{equation}
We refer to $N_{\mathrm{eff},t}$ as the \emph{effective source support}.
It equals one when all absolute mass is assigned to one source and increases
as the mass becomes distributed across sources. This is an
inverse-concentration quantity, not an entropy-based effective support.
We additionally record the largest absolute source share
\begin{equation}
P_{\max,t}
=
\max_{\ell<t}p_{\ell,t}.
\end{equation}

If $A_t=0$, the normalized quantities in
Equations~\ref{eq:absolute_source_weights}--\ref{eq:effective_support} are
undefined and are omitted from the corresponding Persistent summary. All
frozen Persistent targets used for the reported exact-provenance analysis
have nonzero attribution mass, so this rule does not remove any reported
Persistent target.

The direct-provenance analysis uses five frozen target layers per family:
\[
\{8,11,15,18,22\}
\quad\text{for SmolLM2},
\qquad
\{5,8,9,12,16\}
\quad\text{for Gemma}.
\]
Each frozen target-layer summary uses the same 96-sequence evaluation pool
described in Appendix~\ref{app:evaluation_roles}. Target-layer summaries are
averaged with equal weight within a checkpoint. Cross-checkpoint summaries
then treat training seeds 42, 43, and 44 as the replication units and use
Student-$t$ 95\% intervals with two degrees of freedom.

\begin{figure*}[t]
\centering
\includegraphics[width=\textwidth]{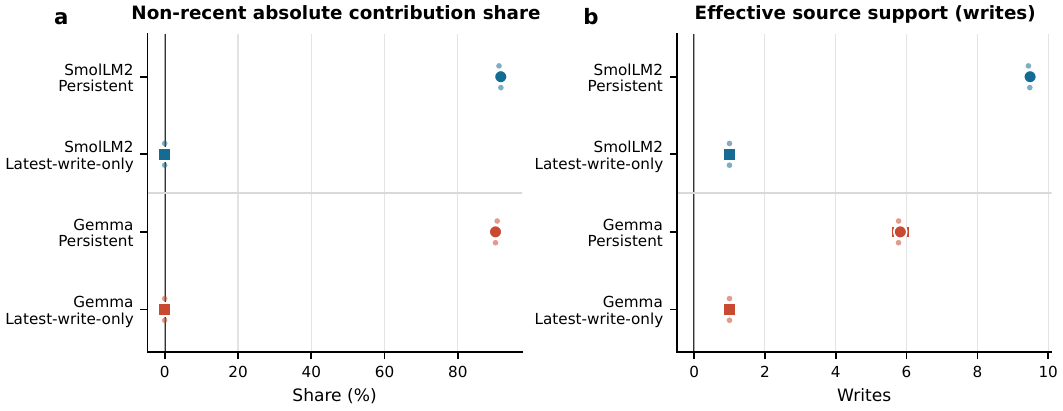}
\caption{
\textbf{Distribution of direct state-mediated routing provenance across
training seeds.}
\textbf{(a)} Fraction of absolute selected-set-margin contribution assigned to
all sources except the most recent.
\textbf{(b)} Effective source support from
Equation~\ref{eq:effective_support}.
For Persistent these statistics are computed from the realized-write ledger.
For latest-write-only they are $0$ and $1$, respectively, under the distinct
structural latest-replacement audit object defined in
Section~\ref{app:lwo_audit}; they are not obtained by applying the Persistent
telescoping ledger to latest-write-only.
Points are three-seed means with Student-$t$ 95\% intervals
($n=3$, df$=2$); $\times$ marks individual training seeds.
}
\label{fig:distributed-history-appendix}
\end{figure*}

Persistent assigns most absolute direct contribution away from the latest
write. The non-recent share is
$91.7\%$ on SmolLM2
(95\% CI $[90.5,92.9]\%$) and
$90.3\%$ on Gemma
($[89.2,91.4]\%$).
The corresponding effective supports are
$9.49$ writes ($[9.36,9.62]$) and
$5.82$ writes ($[5.61,6.04]$).
The largest individual source accounts for only
$20.4\%$ ($[19.8,21.0]\%$) of absolute mass on SmolLM2 and
$31.5\%$ ($[29.5,33.4]\%$) on Gemma.

These quantities use absolute contribution and therefore do not measure net
signed support. As a complementary signed summary, the direct
state-mediated Top-2-boundary contribution,
$\sum_{\ell<t}c_{\ell,t}$, has three-seed means
$0.408$ on SmolLM2 and $0.254$ on Gemma, with 95\% intervals
$[0.239,0.577]$ and $[0.175,0.333]$, respectively.
Large positive and negative source terms may therefore cancel even when
absolute provenance is distributed. The distribution statistics should not be
interpreted as downstream causal effects; those are tested separately in
Appendix~\ref{app:attribution}.

\subsection{Latest-Write-Only Direct-State Audit Object}
\label{app:lwo_audit}

Latest-write-only does not use the Persistent multi-write ledger as its native
audit object. Its recurrence is
\begin{equation}
s_{\ell+1}
=
\operatorname{clip}(u_\ell,-5,5),
\end{equation}
so the state entering a later router directly contains only the most recent
replacement value.

For target layer $t$, let $s_t$ be the state entering the target router and
let $e_2,e_3$ be the clean second-selected and best-excluded experts.
The direct latest-state score assigned to the immediately preceding eligible
source is
\begin{equation}
a_{t,\mathrm{latest}}
=
\left|
\left(
W^t_S[e_2]-W^t_S[e_3]
\right)^\top
s_t
\right|.
\label{eq:lwo_direct_score}
\end{equation}
Every earlier eligible source receives direct native score zero because its
replacement value has already been overwritten. The state and router-weight
slice are converted to float32 before this dot product. Source-score ties are
resolved by descending score and then ascending source-layer index.

The zero scores in this construction have a narrow interpretation: they mean
that an earlier source has zero \emph{direct contribution through the state
currently entering the target router}. They do not imply zero indirect
influence through the ordinary hidden stream or through consequences of
earlier routing decisions.

This distinction is important for Figure~\ref{fig:distributed-history-appendix}.
If one instead mechanically formed
\[
\Delta s_\ell=s_{\ell+1}-s_\ell
\]
under the latest-write-only recurrence, earlier differences would generally
be nonzero and would telescope through cancellation. That is not the audit
object used for the latest-write-only comparison. Under the implemented
structural direct-state definition, the complete direct absolute mass belongs
to the most recent replacement source; hence non-recent share is zero,
effective source support is one, and largest-source share is one by
construction.

Persistent and latest-write-only should therefore be understood as exposing
different native direct-state provenance structures: Persistent retains a
sequence of realized additive state changes, whereas latest-write-only retains
one direct replacement source.

\subsection{Source-Suppression Semantics}
\label{app:state_suppression_semantics}

The full-forward source intervention uses the same hook operation for both
state architectures but has a different recurrence-level interpretation.

For a selected token and source layer $\ell$, a pre-hook first saves the state
$s_\ell$ entering the source block. The source block is then allowed to run
naturally, including its normal routing decision, expert computation, write
proposal, and state update. A post-hook verifies the natural state transition
against the clean recorded transition and then replaces the selected token's
post-source state by the saved pre-source value:
\begin{equation}
\widetilde s_{\ell+1}
=
s_\ell.
\label{eq:source_suppression_common}
\end{equation}
All subsequent model computation is then rerun naturally from this modified
trajectory. Later hidden states, routers, expert selections, write proposals,
states, and output logits are not held fixed.

For Persistent, the natural transition is
\begin{equation}
s_{\ell+1}^{P}
=
\operatorname{clip}(s_\ell+u_\ell,-5,5),
\end{equation}
whereas the intervention uses
\begin{equation}
\widetilde s_{\ell+1}^{P}
=
s_\ell.
\label{eq:persistent_suppression}
\end{equation}
Thus the intervention removes the realized post-clamp state transition at
that source. This formulation is preferable to setting the proposal
$u_\ell$ to zero because saturation can make the realized write
$\Delta s_\ell$ differ from the proposal.

For latest-write-only, the natural transition is
\begin{equation}
s_{\ell+1}^{L}
=
\operatorname{clip}(u_\ell,-5,5),
\end{equation}
while the same hook gives
\begin{equation}
\widetilde s_{\ell+1}^{L}
=
s_\ell.
\label{eq:lwo_suppression}
\end{equation}
This intervention therefore \emph{skips the current replacement and restores
the previous state}; it does not set the current replacement value to zero.
At the next latest-write-only update the explicit state is again replaced by
the newly recomputed proposal. Indirect consequences can nevertheless persist,
because the intervened source may alter subsequent routing, hidden
representations, and therefore later write proposals.

For the scalar audited boundary effect, the identities of the clean $e_2$ and
$e_3$ experts are held fixed so that the same clean margin is evaluated before
and after intervention. Expert selections and other downstream quantities
themselves are allowed to change. Detailed target selection and downstream
metrics are specified in Appendix~\ref{app:attribution}.

\subsection{State Geometry and Scope of the Latest-Write-Only Control}
\label{app:state_geometry}

Persistent and latest-write-only match the routing-state width, write-head
architecture, router input width, parameter count, and analytical active
compute, but they do not induce the same state geometry. Because
\[
u_\ell=\tanh(\cdot),
\]
the latest-write-only state is intrinsically confined to $[-1,1]$ coordinate
wise; its common outer $[-5,5]$ clamp is therefore inactive under exact
arithmetic. Persistent instead accumulates successive proposals and can reach
the $[-5,5]$ clamp.

Consequently, the comparison does not hold state magnitude, variance, or
saturation frequency fixed. It tests persistent accumulation against
replacement under the state distributions learned by those two mechanisms.
We therefore do not interpret the latest-write-only comparison as isolating
accumulation while controlling away every associated geometric or
optimization difference.

An older fixed-checkpoint SmolLM2 diagnostic illustrates the magnitude of
this difference. At source layer 14, the mean state norm was approximately
$42.52$ for Persistent and $9.74$ for latest-write-only, while approximately
$32.96\%$ of Persistent state coordinates were at the clamp boundary.
These values are single-checkpoint diagnostics rather than three-seed
estimates and are not used in the replicated statistical claims of
Section~5. Probe preprocessing, including train-only standardization of native
state features, is specified in Appendix~\ref{app:probe_fitting}.

\section{Attribution, Full-Forward Intervention, and Audit Cost}
\label{app:attribution}

This appendix specifies the fixed target-selection procedure, source-ranking
methods, full-forward intervention metrics, exhaustive ranking diagnostics,
ordinary-MoE post-hoc comparison, and audit-cost measurements underlying the
attribution results in Section~5. Persistent-ledger construction and the
latest-write-only direct-state audit object are defined separately in
Appendix~\ref{app:ledger_distribution}.

\subsection{Target-Sequence and Decision Selection}
\label{app:attribution_targets}

The full-forward intervention study uses 24 fixed final-evaluation sequences
and 96 target routing decisions per model family. Sequence and position
selection are deterministic and outcome independent.

For each of the prose, code, and mathematics domains, the 32 candidate
final-evaluation sequences from Appendix~\ref{app:evaluation_roles} are ranked
by
\begin{equation}
\operatorname{SHA256}\!\left(
\text{salt}\,\Vert\,\texttt{\textbackslash0}\,\Vert\,
\text{architecture}\,\Vert\,\texttt{\textbackslash0}\,\Vert\,
\text{domain}\,\Vert\,\texttt{\textbackslash0}\,\Vert\,
\text{sample\_hash}
\right),
\end{equation}
using salt
\texttt{routing-ledger-confirmatory-e4-sequences-v1}.
Here \texttt{architecture} denotes the model-family identifier,
\texttt{smollm} or \texttt{gemma}, rather than the Persistent,
latest-write-only, or Ordinary architectural variant. Fields are UTF-8
serialized and separated by a single NUL byte. The first eight sequences in
ascending hash order are retained in each domain, giving 24 sequences per
family.

The fixed target layers are
\[
L\in\{15,22\}
\quad\text{for SmolLM2},
\qquad
L\in\{12,16\}
\quad\text{for Gemma}.
\]
Within every selected sequence--target-layer pair, candidate positions are
those with a valid aligned next-token label. They are ranked by
\begin{equation}
\operatorname{SHA256}\!\left(
\text{salt}\,\Vert\,\texttt{\textbackslash0}\,\Vert\,
\text{architecture}\,\Vert\,\texttt{\textbackslash0}\,\Vert\,
\text{sample\_hash}\,\Vert\,\texttt{\textbackslash0}\,\Vert\,
L\,\Vert\,\texttt{\textbackslash0}\,\Vert\,
\text{position}
\right),
\end{equation}
using salt
\texttt{routing-ledger-confirmatory-e4-events-v1}.
The first two positions are retained. Each sequence therefore contributes
four target decisions, for
\[
24\times2\times2=96
\]
targets per family.

The resulting sequence and token identities are reused across matched
architectural variants. Selection does not use routing margins, ledger
contributions, source rankings, intervention effects, output changes, or any
other downstream outcome.

\subsection{Audited Routing Boundary}
\label{app:audited_boundary}

For every retained target decision $i$, let $e_{1,i}$ and $e_{2,i}$ be the
clean first- and second-selected experts and let $e_{3,i}$ be the
clean highest-logit expert excluded from the Top-2 set. The primary audited
quantity is the clean Top-2 membership boundary
\begin{equation}
m_i^{(2,3)}
=
z_{L_i,e_{2,i}}-z_{L_i,e_{3,i}}.
\label{eq:fixed_top2_boundary}
\end{equation}
The identities $e_{2,i}$ and $e_{3,i}$ are fixed from the clean trajectory
when this scalar margin is evaluated after intervention. Routing decisions
themselves are not frozen.

Every MoE layer $\ell<L_i$ is an eligible source. This yields 1,776
source interventions over the 96 SmolLM2 targets and 1,344 over the
96 Gemma targets in the exhaustive SA-MoE boundary audit.

\subsection{SA-MoE Source-Ranking Methods}
\label{app:source_rankings}

All source rankings are computed from the clean trajectory before any
intervention outcome is observed.

\paragraph{Routing ledger.}
The primary ledger score is the magnitude of the exact Persistent
selected-set-margin contribution:
\begin{equation}
q^{\mathrm{ledger}}_{i\ell}
=
\left|
\left(
W^{L_i}_S[e_{2,i}]
-
W^{L_i}_S[e_{3,i}]
\right)^\top
\Delta s_\ell
\right|.
\label{eq:ledger_ranking_score}
\end{equation}

\paragraph{Largest write norm.}
This control ranks sources by
\begin{equation}
q^{\mathrm{norm}}_{i\ell}
=
\|\Delta s_\ell\|_2.
\end{equation}
It tests whether the ledger merely selects unusually large state changes.

\paragraph{Recency.}
The recency baseline ranks later eligible source layers above earlier ones;
its top-ranked source is the greatest $\ell<L_i$.

\paragraph{Gradient $\times$ write.}
The sensitivity-aware score is
\begin{equation}
q^{\mathrm{grad}\times\mathrm{write}}_{i\ell}
=
\left|
\left(
\nabla_{s_{\ell+1}}m_i^{(2,3)}
\right)^\top
\Delta s_\ell
\right|.
\label{eq:gradient_write_score}
\end{equation}
The gradient is evaluated on the clean trajectory with respect to the
post-source state at the same token position. The target expert identities
are the fixed clean $e_2$ and $e_3$. Thus this is a local
clean-trajectory sensitivity score; the realized discrete routing
configuration is not itself replaced by intervention outcomes during ranking.

\paragraph{Matched controls.}
For comparisons against a ledger-selected source, three additional controls
exclude that source. Random selection uses a deterministic SHA-256 ordering
over model family, sample identity, token position, target layer, and
candidate source. Norm matching selects the non-ledger source whose realized
write norm is closest to that of the ledger-selected write. Distance matching
selects deterministically from the same normalized source-depth tercile as
the ledger-selected source. The retained target layers contain enough eligible
sources that every reported depth tercile contains non-ledger candidates.

Unless otherwise specified, deterministic score ties are broken by ascending
source-layer index.

The \emph{ledger-control advantage} used in the replicated persistence
summary is
\begin{equation}
A_i
=
E_{i,\mathrm{ledger}}
-
\frac{1}{3}
\left(
E_{i,\mathrm{random}}
+
E_{i,\mathrm{norm}}
+
E_{i,\mathrm{distance}}
\right),
\label{eq:ledger_control_advantage}
\end{equation}
where $E$ is the absolute fixed-boundary intervention effect defined below.
The checkpoint-level statistic averages $A_i$ over the fixed target set.

\subsection{Full-Forward Source Intervention and Outcome Metrics}
\label{app:intervention_metrics}

State-source suppression follows the architecture-specific semantics in
Appendix~\ref{app:state_suppression_semantics}. In both Persistent and
latest-write-only, the selected token's natural post-source state is replaced
by its pre-source state. The model then continues normally from the modified
trajectory. Later hidden representations, writes, routing decisions, experts,
and output logits may all change.

For target $i$ and source $\ell$, define the primary absolute fixed-boundary
effect
\begin{equation}
E_{i\ell}
=
\left|
m_{i,\mathrm{intervene}(\ell)}^{(2,3)}
-
m_{i,\mathrm{clean}}^{(2,3)}
\right|.
\label{eq:intervention_effect}
\end{equation}
We also retain the corresponding signed difference.

For clean and intervened Top-2 expert sets $S$ and $\widetilde S$, the
routing-set distance is the Jaccard distance
\begin{equation}
d_J(S,\widetilde S)
=
1-
\frac{|S\cap\widetilde S|}
     {|S\cup\widetilde S|}.
\label{eq:jaccard_distance}
\end{equation}
\emph{Target Top-2 Jaccard distance} applies
Equation~\ref{eq:jaccard_distance} at the audited target router.
\emph{Target Top-1 change} is the indicator that the clean and intervened
Top-1 target experts differ. The reported \emph{downstream Top-2 divergence}
averages the corresponding Top-2 Jaccard distance over routing decisions
strictly downstream of the audited target for the selected token.

At the selected token's language-model output, let
$p_{\mathrm{clean}}$ and $p_{\mathrm{int}}$ denote the clean and intervened
softmax distributions. Output KL is reported in the clean-to-intervention
direction,
\begin{equation}
D_{\mathrm{KL}}
\!\left(
p_{\mathrm{clean}}
\Vert
p_{\mathrm{int}}
\right)
=
\sum_v
p_{\mathrm{clean}}(v)
\log
\frac{p_{\mathrm{clean}}(v)}
     {p_{\mathrm{int}}(v)},
\label{eq:output_kl}
\end{equation}
using natural logarithms. Output Top-1 change records whether the most likely
output token changes.

For the aligned next-token label $y_i$, the NLL effect is
\begin{equation}
\Delta\mathrm{NLL}_i
=
-\log p_{\mathrm{int}}(y_i)
+
\log p_{\mathrm{clean}}(y_i),
\label{eq:nll_change}
\end{equation}
so positive values indicate increased loss after intervention.

Within a checkpoint, target-level quantities are first grouped by their
24 source sequences. Confidence intervals use 2,000 paired sequence-level
bootstrap resamples; multiple target layers and positions from one source
sequence are not treated as independent replicates. For architecture-level
comparisons across checkpoints, training seed is the replication unit and
95\% Student-$t$ intervals use $n=3$ and two degrees of freedom.

\subsection{Exhaustive Source-Ranking Metrics}
\label{app:ranking_metrics}

The exhaustive seed-42 audit intervenes on every eligible source for every
fixed target. Let
\begin{equation}
E_i^\star
=
\max_{\ell<L_i} E_{i\ell}
\end{equation}
denote the oracle fixed-boundary effect, and let
$\widehat\ell_{iq}$ be the top-ranked source under method $q$.

Raw oracle regret is
\begin{equation}
R_{iq}
=
E_i^\star
-
E_{i,\widehat\ell_{iq}},
\end{equation}
and normalized oracle regret is
\begin{equation}
\widetilde R_{iq}
=
\frac{
E_i^\star-E_{i,\widehat\ell_{iq}}
}{
E_i^\star
}.
\label{eq:normalized_regret}
\end{equation}
Normalized regret is defined for targets with positive oracle effect. A
zero-oracle target would be excluded from this normalized quantity and
reported separately rather than used as a numerical denominator.

Top-source accuracy is one when the method-selected source belongs to the
set of oracle-maximal sources and zero otherwise; hence ties among genuinely
oracle-maximal sources all count as correct. Within-target Spearman is the
rank correlation between the complete clean-trajectory source-score vector
and the complete intervention-effect vector
$\{E_{i\ell}\}_{\ell<L_i}$. It is reported only when both rankings are
defined; a constant or unavailable score vector contributes no Spearman value
rather than an artificial zero correlation.

\subsection{Fixed-Checkpoint SA-MoE Ranking Diagnostic}
\label{app:samoe_ranking}

Table~\ref{tab:samoe_ranking} reports the exhaustive Persistent seed-42
diagnostic. These are sequence-bootstrap summaries within one trained
checkpoint, not three-seed architecture-level estimates.

\begin{table*}[t]
\centering
\small
\setlength{\tabcolsep}{5pt}
\caption{Exhaustive full-forward ranking quality on the fixed Persistent
seed-42 checkpoints. Entries are means with 95\% paired sequence-bootstrap
intervals over 24 source sequences and 96 targets. Higher Spearman and
top-source accuracy are better; lower normalized oracle regret is better.
Random supplies a selected-source baseline but no complete deterministic
score vector for the SA-MoE Spearman calculation.}
\label{tab:samoe_ranking}
\begin{tabular}{llccc}
\toprule
Family & Method &
Spearman &
Top-source acc. &
Normalized regret \\
\midrule
SmolLM2
& Random
& --- &
$0.073\,[0.021,0.125]$ &
$0.706\,[0.646,0.758]$ \\
& Recency
& $0.150\,[0.092,0.205]$ &
$0.188\,[0.115,0.271]$ &
$0.505\,[0.430,0.580]$ \\
& Largest write norm
& $0.187\,[0.132,0.241]$ &
$0.135\,[0.062,0.219]$ &
$0.499\,[0.437,0.562]$ \\
& Ledger
& $0.369\,[0.319,0.421]$ &
$0.250\,[0.156,0.344]$ &
$0.362\,[0.307,0.417]$ \\
& Gradient $\times$ write
& $0.769\,[0.723,0.807]$ &
$0.771\,[0.688,0.844]$ &
$0.055\,[0.032,0.081]$ \\
\midrule
Gemma
& Random
& --- &
$0.031\,[0.000,0.062]$ &
$0.742\,[0.687,0.792]$ \\
& Recency
& $-0.154\,[-0.201,-0.106]$ &
$0.125\,[0.062,0.198]$ &
$0.629\,[0.573,0.687]$ \\
& Largest write norm
& $0.341\,[0.293,0.389]$ &
$0.521\,[0.417,0.625]$ &
$0.267\,[0.198,0.331]$ \\
& Ledger
& $0.823\,[0.779,0.860]$ &
$0.802\,[0.729,0.865]$ &
$0.058\,[0.033,0.086]$ \\
& Gradient $\times$ write
& $0.890\,[0.871,0.907]$ &
$0.844\,[0.771,0.917]$ &
$0.030\,[0.016,0.046]$ \\
\bottomrule
\end{tabular}
\end{table*}

The ledger is therefore informative about full-forward suppression effects
on its audited boundary, but exact forward contribution is not the same
objective as intervention sensitivity. In this fixed-checkpoint diagnostic,
gradient$\times$write provides the stronger ranking under all three displayed
metrics. We consequently use the ledger as a provenance score, not as a claim
of optimal causal-source ranking.

\subsection{Ordinary-MoE Post-Hoc Attribution}
\label{app:li_attribution}

We separately apply the cross-layer routing-attribution method of
\citet{li2026understanding} to the matched Ordinary MoEs. This analysis uses
ordinary-MoE residual components as its audit objects rather than SA-MoE
state writes. Raw component-intervention magnitudes are therefore not pooled
with or directly compared against SA-MoE write-intervention magnitudes.

The implementation was checked against the authors' public reference code at
commit
\texttt{fe2366a914f1da426ab7ec4d903015e229beaa72}.
The upstream decomposition helpers were directly parity-tested. The maximum
absolute discrepancies were $1.19\times10^{-7}$ for the default difference
breakdown and $2.98\times10^{-8}$ for the RMSNorm/TAM helper. The Gemma
adapter uses Gemma's effective RMSNorm scale, including the implementation's
$(1+\text{parameter})$ convention.

We retain two Li-style scores. \emph{Li variance} uses the reference
expert-score variance criterion. \emph{Li Top-2 boundary} uses the same
decomposed ordinary-MoE components but adapts the final score to the fixed
clean $e_2$-versus-$e_3$ routing boundary used in our intervention benchmark.
We additionally evaluate largest-component norm, recency, deterministic
random ranking, and a gradient$\times$component sensitivity score.

As a numerical validation of the decomposition, the maximum reconstructed
router-input error is $2.384\times10^{-7}$ on SmolLM2 and
$1.907\times10^{-6}$ on Gemma in the frozen validation run; maximum router-logit
reconstruction error is zero for both.

Unlike the single-checkpoint SA-MoE exhaustive diagnostic above, the current
ordinary-MoE ranking summaries are available for training seeds 42, 43, and
44. Table~\ref{tab:ordinary_li_ranking} therefore reports three-seed means
with Student-$t$ intervals. The column labeled ``regret'' preserves the
ordinary-MoE post-hoc artifact's reported oracle-regret statistic; it should
not be numerically pooled with SA-MoE intervention magnitudes.

\begin{table*}[t]
\centering
\small
\setlength{\tabcolsep}{4.5pt}
\caption{Ordinary-MoE attribution ranking across training seeds 42/43/44.
Entries are three-seed means with 95\% Student-$t$ intervals
($n=3$, df$=2$). Higher Spearman and top-source accuracy are better; lower
regret is better.}
\label{tab:ordinary_li_ranking}
\begin{tabular}{llccc}
\toprule
Family & Method & Spearman & Top-source acc. & Regret \\
\midrule
SmolLM2
& Random
& $-0.030\,[-0.046,-0.014]$
& $0.007\,[-0.023,0.037]$
& $0.916\,[0.905,0.928]$ \\
& Recency
& $0.125\,[0.089,0.161]$
& $0.090\,[0.051,0.130]$
& $0.683\,[0.616,0.750]$ \\
& Largest component norm
& $0.662\,[0.638,0.687]$
& $0.319\,[0.240,0.398]$
& $0.503\,[0.429,0.578]$ \\
& Li variance
& $0.653\,[0.634,0.672]$
& $0.090\,[0.060,0.120]$
& $0.698\,[0.592,0.803]$ \\
& Li Top-2 boundary
& $0.570\,[0.542,0.599]$
& $0.212\,[0.172,0.251]$
& $0.581\,[0.510,0.652]$ \\
& Gradient $\times$ component
& $0.721\,[0.670,0.771]$
& $0.326\,[0.154,0.499]$
& $0.404\,[0.278,0.530]$ \\
\midrule
Gemma
& Random
& $-0.068\,[-0.130,-0.007]$
& $0.035\,[-0.005,0.074]$
& $0.853\,[0.828,0.879]$ \\
& Recency
& $0.231\,[0.196,0.266]$
& $0.330\,[0.197,0.463]$
& $0.373\,[0.356,0.390]$ \\
& Largest component norm
& $0.600\,[0.564,0.637]$
& $0.260\,[0.167,0.354]$
& $0.455\,[0.355,0.556]$ \\
& Li variance
& $0.672\,[0.651,0.692]$
& $0.375\,[0.218,0.532]$
& $0.347\,[0.239,0.455]$ \\
& Li Top-2 boundary
& $0.620\,[0.572,0.668]$
& $0.389\,[0.223,0.555]$
& $0.300\,[0.207,0.393]$ \\
& Gradient $\times$ component
& $0.782\,[0.761,0.803]$
& $0.521\,[0.384,0.658]$
& $0.207\,[0.176,0.238]$ \\
\bottomrule
\end{tabular}
\end{table*}

These results confirm that cross-layer routing attribution does not require
an explicit routing state. Li-style post-hoc attribution recovers meaningful
source rankings in the Ordinary MoEs, while gradient-based sensitivity is
again competitive or stronger on the intervention-ranking objective. The
distinction claimed for SA-MoE is therefore the availability and semantics of
its routing-specific forward audit object, not exclusive access to
cross-layer routing information.

\subsection{Audit Cost Under a Common Capture Policy}
\label{app:attribution_costs}

Audit cost depends on what is retained during the clean forward pass. We
therefore do not equate ``post-hoc'' with mandatory model replay.

For SA-MoE, retaining every bf16 realized 128-dimensional write from a
length-2048 forward pass requires 15.00 MiB for SmolLM2 and 9.00 MiB for
Gemma. Retaining the clean Top-3 expert identities and logits in addition
raises these routing-audit traces to approximately 15.70 MiB and 9.42 MiB,
respectively. The live 128-dimensional state alone is 0.50 MiB for a
length-2048 sequence.

If this write trace is retained, later ledger contribution queries require
neither another model forward pass nor a target-dependent backward pass.
Likewise, if the ordinary-MoE residual components required by the Li
decomposition are retained during the clean forward, Li-style fixed-denominator
scores can be formed without model replay or backward propagation. The frozen
cost evidence does not provide an equivalently defined serialized retained-byte
footprint for that component trace, so we do not report an asymmetric byte
comparison.

Gradient$\times$write and gradient$\times$component additionally require
target-dependent gradient information unless a differentiable clean graph is
itself retained, which has a different memory trade-off.

Table~\ref{tab:audit_cost_pilot} gives phase-isolated pilot measurements on a
single NVIDIA GH200 480GB GPU. Each pilot uses one frozen sequence, batch size
8, and four target decisions. Model loading, immutable-input validation,
ranking aggregation, and causal-oracle interventions are excluded from the
displayed extraction phases. These timings are diagnostic measurements rather
than the scientific 96-target ranking experiment.

\begin{table*}[t]
\centering
\small
\setlength{\tabcolsep}{4pt}
\caption{Phase-isolated attribution-cost pilot. ``Capture'' is one clean
forward with the minimal representation required by the corresponding audit
object. Peak is maximum allocated accelerator memory during that capture.
For the native ledger, contribution materialization occurs during capture and
requires zero incremental model passes. Li score extraction operates on
captured components and likewise uses zero incremental forwards or backwards.
Gradient extraction is shown separately because it is target dependent.}
\label{tab:audit_cost_pilot}
\begin{tabular}{llrrrr}
\toprule
Family & Audit object &
Capture (s) &
Capture peak (GiB) &
Post-capture score (s) &
Gradient extraction (s) \\
\midrule
SmolLM2
& Persistent ledger
& 0.249 & 3.96 & included & 3.205 \\
& Ordinary Li components
& 0.752 & 2.85 & 0.152 & 1.146 \\
\midrule
Gemma
& Persistent ledger
& 0.428 & 10.88 & included & 1.260 \\
& Ordinary Li components
& 0.852 & 10.70 & 0.080 & 1.319 \\
\bottomrule
\end{tabular}
\end{table*}

The pilot environment uses Python 3.12.14, PyTorch
\texttt{2.10.0+cu130}, and Transformers 5.1.0. The Persistent pilot contains
74 eligible source interventions over four SmolLM2 targets and 56 over four
Gemma targets; the ordinary-MoE component audit contains 222 and 168 source
components, respectively. Gradient extraction uses four autograd calls for
the four targets.

These measurements establish an operational distinction but not a universal
runtime ordering. A retained SA-MoE write trace supports later algebraic
ledger queries directly; a sufficiently rich retained ordinary-MoE component
trace can likewise avoid replay for Li-style queries. Retention policy,
component granularity, target count, and whether gradients are required all
change the relevant cost.

\subsection{Replicated Downstream Consequences}
\label{app:downstream_seed_results}

The main downstream results use the ledger-selected source for Persistent and
the corresponding native latest-state-selected source for latest-write-only.
Table~\ref{tab:replicated_downstream} reports the three-seed means underlying
the main-text intervention figure. Complete seed values are additionally
collected in Appendix~\ref{app:seed_results}.

\begin{table*}[t]
\centering
\small
\setlength{\tabcolsep}{4.5pt}
\caption{Full-forward source-suppression outcomes across training seeds.
Entries are three-seed means with 95\% Student-$t$ intervals
($n=3$, df$=2$). The P--LWO row is the paired Persistent-minus-latest-write-only
contrast.}
\label{tab:replicated_downstream}
\begin{tabular}{llccc}
\toprule
Family & Architecture &
Target $|\Delta m^{(2,3)}|$ &
Downstream Top-2 divergence &
Output KL \\
\midrule
SmolLM2
& Persistent
& $0.276\,[0.156,0.396]$
& $0.080\,[0.040,0.120]$
& $0.00155\,[0.00115,0.00195]$ \\
& Latest-write-only
& $0.180\,[-0.057,0.417]$
& $0.030\,[0.010,0.051]$
& $0.00159\,[0.00107,0.00211]$ \\
& P--LWO
& $0.096\,[-0.179,0.371]$
& $0.050\,[-0.009,0.108]$
& $-0.00004\,[-0.00064,0.00056]$ \\
\midrule
Gemma
& Persistent
& $0.426\,[0.338,0.515]$
& $0.195\,[0.126,0.264]$
& $0.01615\,[0.00521,0.02709]$ \\
& Latest-write-only
& $0.279\,[0.151,0.407]$
& $0.049\,[0.003,0.095]$
& $0.00609\,[0.00063,0.01156]$ \\
& P--LWO
& $0.147\,[-0.034,0.328]$
& $0.146\,[0.041,0.251]$
& $0.01006\,[-0.00510,0.02521]$ \\
\bottomrule
\end{tabular}
\end{table*}

For the ledger-control advantage of
Equation~\ref{eq:ledger_control_advantage}, Persistent has a three-seed mean
of $0.140$ margin units on SmolLM2
(95\% CI $[0.009,0.270]$) and $0.328$ on Gemma
($[0.235,0.420]$). The corresponding latest-write-only values are
$0.083$ ($[-0.141,0.306]$) and
$0.215$ ($[0.087,0.342]$). The paired Persistent-minus-latest-write-only
contrasts are $0.057$ ($[-0.213,0.327]$) and
$0.113$ ($[-0.073,0.299]$), respectively.

Thus, ledger-selected Persistent writes have replicated effects on the
audited target margin, later routing, and output distribution. Persistence
does not, however, produce a uniformly resolved increase over
latest-write-only for every intervention outcome, and the fixed-checkpoint
ranking analysis shows that gradient-based sensitivity can outperform exact
forward provenance when the objective is specifically to identify the source
with the largest full-forward perturbation.

\section{Complete Per-Seed Results and Analysis Scope}
\label{app:seed_results}

This appendix collects the training-seed values underlying the replicated
results in the main text and Appendices~\ref{app:evaluation}--\ref{app:attribution}.
Unless stated otherwise, training seeds $42$, $43$, and $44$ are the
replication units. Cross-seed intervals reported elsewhere in the paper use
Student-$t$ 95\% intervals with two degrees of freedom. Within-checkpoint
bootstrap samples are never treated as additional training replicates.

\subsection{Model Quality and Downstream Capability}
\label{app:seed_quality}

Table~\ref{tab:seed_quality} reports the seed-level quality contrasts underlying
Section~5.1 and Appendix~\ref{app:capability}. PPL is shown as the ratio to the
matched Ordinary MoE; task columns are absolute percentage-point changes in
the primary task metric. Thus a PPL ratio above one indicates higher
perplexity, while a positive task value indicates higher accuracy.

\begin{table*}[t]
\centering
\small
\setlength{\tabcolsep}{4.5pt}
\caption{Seed-level quality changes relative to matched Ordinary MoE.
HellaSwag, ARC-Easy, and PIQA use normalized accuracy; WinoGrande uses raw
accuracy. Macro is their unweighted mean.}
\label{tab:seed_quality}
\begin{tabular}{lllrrrrrr}
\toprule
Family & Architecture & Seed &
PPL ratio &
$\Delta$Hella &
$\Delta$ARC-E &
$\Delta$PIQA &
$\Delta$Wino &
$\Delta$Macro \\
\midrule
SmolLM2
& Latest-write-only & 42 & 1.00361 & $-0.159$ & $-0.505$ & $-0.054$ & $+0.158$ & $-0.140$ \\
&                   & 43 & 1.00557 & $+0.309$ & $-0.715$ & $+0.054$ & $+1.105$ & $+0.188$ \\
&                   & 44 & 1.00625 & $+0.259$ & $+0.126$ & $+0.871$ & $-1.736$ & $-0.120$ \\
& Persistent        & 42 & 1.00213 & $-0.020$ & $-0.210$ & $+0.326$ & $-0.158$ & $-0.015$ \\
&                   & 43 & 1.00187 & $+0.020$ & $-1.389$ & $+0.381$ & $+0.631$ & $-0.089$ \\
&                   & 44 & 1.00161 & $+0.030$ & $+0.084$ & $+0.381$ & $-0.710$ & $-0.054$ \\
\midrule
Gemma
& Latest-write-only & 42 & 1.00077 & $+0.149$ & $-2.315$ & $-0.925$ & $+0.947$ & $-0.536$ \\
&                   & 43 & 1.00432 & $-0.518$ & $-0.673$ & $-0.109$ & $-1.894$ & $-0.799$ \\
&                   & 44 & 1.00399 & $-0.797$ & $-0.295$ & $+0.054$ & $+0.474$ & $-0.141$ \\
& Persistent        & 42 & 1.00096 & $-0.060$ & $-1.305$ & $-0.435$ & $+1.105$ & $-0.174$ \\
&                   & 43 & 0.99928 & $-0.558$ & $+0.505$ & $-0.762$ & $-2.210$ & $-0.756$ \\
&                   & 44 & 0.99787 & $-0.717$ & $-0.295$ & $+1.415$ & $+1.421$ & $+0.456$ \\
\bottomrule
\end{tabular}
\end{table*}

Positive-ratio quantities such as PPL are analyzed across seeds on the
natural-log scale before transforming the mean and confidence interval back to
the ratio scale. Accuracy differences and other additive quantities use the
ordinary arithmetic scale.

\subsection{State Use and Persistence Endpoints}
\label{app:seed_state_persistence}

Table~\ref{tab:seed_dose_response} gives the seed-level Persistent
dose-response values underlying Figure~\ref{fig:state-use-persistence}a.
The no-op endpoint $\lambda=1$ is exactly zero by construction.

\begin{table}[t]
\centering
\small
\setlength{\tabcolsep}{5pt}
\caption{Persistent selected-Top-2 set-change rate under local state
attenuation, in percent. The intervention holds the natural hidden
representation and state fixed and changes only the current router's direct
state contribution.}
\label{tab:seed_dose_response}
\begin{tabular}{lrrrr}
\toprule
Family & $\lambda$ & Seed 42 & Seed 43 & Seed 44 \\
\midrule
SmolLM2 & 0.75 & 11.89 & 12.34 & 11.37 \\
        & 0.50 & 28.88 & 30.22 & 27.10 \\
        & 0.25 & 54.25 & 56.48 & 51.61 \\
        & 0    & 88.16 & 88.61 & 86.17 \\
\midrule
Gemma   & 0.75 & 17.88 & 17.03 & 15.94 \\
        & 0.50 & 34.67 & 35.76 & 33.91 \\
        & 0.25 & 52.08 & 53.47 & 51.44 \\
        & 0    & 69.12 & 70.93 & 69.67 \\
\bottomrule
\end{tabular}
\end{table}

At full suppression, the corresponding latest-write-only rates are
$(71.77,72.22,71.08)\%$ for SmolLM2 and
$(58.66,57.03,58.46)\%$ for Gemma.

Table~\ref{tab:seed_persistence_endpoints} gives the exact seed-level
contrasts used in Figure~\ref{fig:state-use-persistence}b. These are the
frozen \emph{persistence-endpoint} reductions. They should not be substituted
for the separately aggregated Figure~\ref{fig:ledger-downstream} quantities
reported later in this appendix.

\begin{table*}[t]
\centering
\small
\setlength{\tabcolsep}{5pt}
\caption{Seed-level Persistent-minus-latest-write-only persistence endpoints.
Long-horizon accessibility is in percentage points; the two intervention
endpoints are in target-margin units. The final column gives the three-seed
mean and Student-$t$ 95\% interval.}
\label{tab:seed_persistence_endpoints}
\begin{tabular}{llrrrr}
\toprule
Family & Endpoint & Seed 42 & Seed 43 & Seed 44 & Mean [95\% CI] \\
\midrule
SmolLM2
& Long-horizon accessibility (pp)
& 21.97 & 19.33 & 16.99
& $19.43\,[13.24,25.62]$ \\
& Ledger--control advantage
& 0.143 & 0.060 & $-0.009$
& $0.065\,[-0.124,0.253]$ \\
& Ledger intervention effect
& 0.174 & 0.124 & 0.016
& $0.105\,[-0.096,0.305]$ \\
\midrule
Gemma
& Long-horizon accessibility (pp)
& 9.68 & 15.09 & 11.75
& $12.17\,[5.40,18.94]$ \\
& Ledger--control advantage
& 0.046 & 0.234 & 0.040
& $0.106\,[-0.167,0.380]$ \\
& Ledger intervention effect
& 0.059 & 0.257 & 0.086
& $0.134\,[-0.134,0.402]$ \\
\bottomrule
\end{tabular}
\end{table*}

The long-horizon contrast is positive in all three training seeds for both
model families. The two intervention-derived persistence contrasts have
positive three-seed means but are less uniform and have intervals spanning
zero.

\subsection{Future-Routing Probe Curves}
\label{app:seed_probe_values}

Table~\ref{tab:seed_probe_curves} gives the complete seed-level numerical
record for the three representations used in the replicated main
future-routing comparison: the Persistent native state, latest-write-only
native state, and supervised 128-dimensional subspace of the Persistent
model's ordinary hidden stream. Values are transition-only accuracy,
equal-pair averaged across all six frozen source--target pairs as defined in
Appendix~\ref{app:probe_aggregation}.

\begin{table*}[t]
\centering
\small
\setlength{\tabcolsep}{6pt}
\caption{Seed-level transition-only future-routing accuracy (\%) underlying
Figure~\ref{fig:probe-curves}. ``P hidden-128'' denotes the supervised
128-dimensional representation learned from the Persistent model's hidden
stream.}
\label{tab:seed_probe_curves}
\begin{tabular}{lllrrr}
\toprule
Family & Budget & Representation & Seed 42 & Seed 43 & Seed 44 \\
\midrule
SmolLM2
& 10k  & Persistent native & 79.28 & 78.37 & 80.18 \\
&      & Latest-write-only native & 57.90 & 56.24 & 56.48 \\
&      & P hidden-128 & 76.56 & 77.93 & 78.81 \\
& 100k & Persistent native & 81.18 & 79.88 & 82.12 \\
&      & Latest-write-only native & 59.50 & 57.76 & 58.40 \\
&      & P hidden-128 & 81.52 & 82.20 & 82.92 \\
& 1M   & Persistent native & 81.45 & 79.96 & 82.33 \\
&      & Latest-write-only native & 59.68 & 57.96 & 58.56 \\
&      & P hidden-128 & 82.94 & 83.18 & 83.98 \\
\midrule
Gemma
& 10k  & Persistent native & 73.91 & 75.50 & 75.45 \\
&      & Latest-write-only native & 63.47 & 64.05 & 66.76 \\
&      & P hidden-128 & 75.73 & 79.61 & 79.22 \\
& 100k & Persistent native & 76.06 & 78.24 & 77.16 \\
&      & Latest-write-only native & 65.81 & 65.68 & 68.82 \\
&      & P hidden-128 & 81.24 & 83.19 & 83.68 \\
& 1M   & Persistent native & 76.48 & 78.33 & 77.41 \\
&      & Latest-write-only native & 66.23 & 65.77 & 68.96 \\
&      & P hidden-128 & 82.70 & 84.69 & 84.91 \\
\bottomrule
\end{tabular}
\end{table*}

At the $10^6$-example budget, the within-seed Persistent-minus-latest-write-only
native-state gaps are
\[
(21.78,\;22.00,\;23.77)\ \text{pp}
\]
for SmolLM2 and
\[
(10.25,\;12.55,\;8.45)\ \text{pp}
\]
for Gemma. Their three-seed means are the 22.5 pp and 10.4 pp values reported
in Section~\ref{sec:results_probes}. These six-pair values are distinct from
the four-pair long-horizon endpoint in
Table~\ref{tab:seed_persistence_endpoints}.

\subsection{Distributed Persistent History}
\label{app:seed_history}

Table~\ref{tab:seed_history} collects the seed-level Persistent
direct-provenance summaries underlying
Appendix~\ref{app:history_metrics}. Latest-write-only has non-recent share
zero, effective support one, and largest-source share one by construction
under the structural direct-state audit object of
Appendix~\ref{app:lwo_audit}.

\begin{table*}[t]
\centering
\small
\setlength{\tabcolsep}{5pt}
\caption{Seed-level Persistent direct-provenance summaries. Contribution-share
columns use absolute selected-set-margin contribution. Signed margin is the
net direct state-mediated selected-set-margin contribution.}
\label{tab:seed_history}
\begin{tabular}{llrrrr}
\toprule
Family & Metric & Seed 42 & Seed 43 & Seed 44 & Mean [95\% CI] \\
\midrule
SmolLM2
& Non-recent absolute share (\%)
& 91.24 & 92.20 & 91.76
& $91.73\,[90.55,92.92]$ \\
& Effective source support
& 9.444 & 9.544 & 9.482
& $9.490\,[9.364,9.616]$ \\
& Largest absolute source share (\%)
& 20.57 & 20.10 & 20.47
& $20.38\,[19.77,21.00]$ \\
& Signed direct margin
& 0.478 & 0.343 & 0.403
& $0.408\,[0.239,0.577]$ \\
\midrule
Gemma
& Non-recent absolute share (\%)
& 90.75 & 89.84 & 90.28
& $90.29\,[89.15,91.43]$ \\
& Effective source support
& 5.776 & 5.923 & 5.774
& $5.824\,[5.613,6.036]$ \\
& Largest absolute source share (\%)
& 32.12 & 30.58 & 31.70
& $31.47\,[29.49,33.45]$ \\
& Signed direct margin
& 0.269 & 0.276 & 0.218
& $0.254\,[0.175,0.333]$ \\
\bottomrule
\end{tabular}
\end{table*}

\subsection{Ordinary-MoE Post-Hoc Attribution Across Seeds}
\label{app:seed_li_results}

For completeness, Table~\ref{tab:seed_li_results} gives the training-seed
values underlying the principal Ordinary-MoE attribution comparison in
Section~5.5 and Appendix~\ref{app:li_attribution}. These quantities concern
ordinary-MoE residual components and are not pooled with SA-MoE write
interventions.

\begin{table*}[t]
\centering
\small
\setlength{\tabcolsep}{5pt}
\caption{Seed-level Ordinary-MoE attribution-ranking results for the
Li Top-2-boundary adaptation and gradient$\times$component. Higher Spearman
and top-source accuracy are better; lower regret is better.}
\label{tab:seed_li_results}
\begin{tabular}{lllrrr}
\toprule
Family & Method & Metric & Seed 42 & Seed 43 & Seed 44 \\
\midrule
SmolLM2
& Li Top-2 boundary & Spearman & 0.577 & 0.557 & 0.576 \\
&                   & Top-source accuracy & 0.208 & 0.198 & 0.229 \\
&                   & Regret & 0.575 & 0.612 & 0.555 \\
& Gradient $\times$ component
& Spearman & 0.731 & 0.697 & 0.734 \\
&                   & Top-source accuracy & 0.281 & 0.292 & 0.406 \\
&                   & Regret & 0.413 & 0.450 & 0.350 \\
\midrule
Gemma
& Li Top-2 boundary & Spearman & 0.600 & 0.638 & 0.623 \\
&                   & Top-source accuracy & 0.438 & 0.312 & 0.417 \\
&                   & Regret & 0.322 & 0.321 & 0.257 \\
& Gradient $\times$ component
& Spearman & 0.774 & 0.791 & 0.780 \\
&                   & Top-source accuracy & 0.479 & 0.500 & 0.583 \\
&                   & Regret & 0.221 & 0.204 & 0.197 \\
\bottomrule
\end{tabular}
\end{table*}

\subsection{Full-Forward Downstream Consequences}
\label{app:seed_downstream}

Table~\ref{tab:seed_downstream} gives the exact training-seed values underlying
Figure~\ref{fig:ledger-downstream} and
Table~\ref{tab:replicated_downstream}. ``P--LWO'' denotes the within-seed
Persistent-minus-latest-write-only contrast.

\begin{table*}[t]
\centering
\small
\setlength{\tabcolsep}{5pt}
\caption{Seed-level full-forward source-suppression outcomes.
Target $|\Delta m|$ is the absolute change in the fixed clean Top-2 boundary;
downstream Top-2 is mean Jaccard distance; KL is
$D_{\mathrm{KL}}(p_{\mathrm{clean}}\Vert p_{\mathrm{int}})$.}
\label{tab:seed_downstream}
\begin{tabular}{lllrrr}
\toprule
Family & Architecture / contrast & Metric & Seed 42 & Seed 43 & Seed 44 \\
\midrule
SmolLM2
& Persistent
& Target $|\Delta m|$ & 0.3289 & 0.2344 & 0.2641 \\
& & Downstream Top-2 & 0.0808 & 0.0958 & 0.0638 \\
& & Output KL & 0.001386 & 0.001709 & 0.001556 \\
& Latest-write-only
& Target $|\Delta m|$ & 0.1365 & 0.1138 & 0.2891 \\
& & Downstream Top-2 & 0.0257 & 0.0257 & 0.0400 \\
& & Output KL & 0.001442 & 0.001504 & 0.001832 \\
& P--LWO
& Target $|\Delta m|$ & 0.1924 & 0.1206 & $-0.0250$ \\
& & Downstream Top-2 & 0.0551 & 0.0701 & 0.0238 \\
& & Output KL & $-0.000056$ & 0.000206 & $-0.000276$ \\
\midrule
Gemma
& Persistent
& Target $|\Delta m|$ & 0.3850 & 0.4488 & 0.4446 \\
& & Downstream Top-2 & 0.2222 & 0.1968 & 0.1667 \\
& & Output KL & 0.013094 & 0.021198 & 0.014161 \\
& Latest-write-only
& Target $|\Delta m|$ & 0.2982 & 0.2206 & 0.3181 \\
& & Downstream Top-2 & 0.0457 & 0.0324 & 0.0689 \\
& & Output KL & 0.008626 & 0.004632 & 0.005024 \\
& P--LWO
& Target $|\Delta m|$ & 0.0868 & 0.2282 & 0.1266 \\
& & Downstream Top-2 & 0.1765 & 0.1644 & 0.0978 \\
& & Output KL & 0.004468 & 0.016566 & 0.009137 \\
\bottomrule
\end{tabular}
\end{table*}

The seed-level record makes the qualification in Section~5.5 explicit.
Persistent ledger-selected suppression has nonzero mean downstream effects in
both families, but Persistent-minus-latest-write-only differences are not
uniformly positive for every outcome and training seed.

\subsection{Scope of Statistical Claims}
\label{app:analysis_scope}

Table~\ref{tab:analysis_scope} summarizes the replication and uncertainty unit
for each class of analysis. This distinction is important because several
supplementary forensic analyses intentionally use one frozen checkpoint and
should not be interpreted as three-seed architecture-level evidence.

\begin{table*}[t]
\centering
\small
\setlength{\tabcolsep}{5pt}
\caption{Scope of reported analyses. Training seed is used as the replication
unit only where three independently trained checkpoints are available.
Sequence-bootstrap intervals characterize within-checkpoint evaluation
variation and are not substitutes for training replication.}
\label{tab:analysis_scope}
\begin{tabular}{p{0.25\textwidth}p{0.22\textwidth}p{0.23\textwidth}p{0.22\textwidth}}
\toprule
Analysis & Checkpoint scope & Uncertainty / replication unit & Interpretation \\
\midrule
Model quality and capability
& Seeds 42/43/44 for each family and architecture
& Training seed; Student-$t$ ($df=2$). PPL analyzed on log-ratio scale.
& Replicated architecture-level result. \\

Local state-suppression dose response
& Persistent and latest-write-only, seeds 42/43/44
& Training seed; Student-$t$ ($df=2$)
& Replicated local router-dependence result. \\

Long-horizon persistence endpoint
& Persistent and latest-write-only, seeds 42/43/44
& Paired training seed; Student-$t$ ($df=2$)
& Primary replicated persistence result. \\

Main future-routing probe curves
& Seeds 42/43/44
& Training seed after within-checkpoint fit/pair reduction; Student-$t$ ($df=2$)
& Replicated representation-accessibility result. \\

Additional hidden-state probe controls
& Supplementary frozen representation analyses
& Within-checkpoint probe evaluation unless explicitly reported across seeds
& Representation diagnostic; not used as an independent training-replication claim. \\

Distributed-history provenance
& Persistent and latest-write-only, seeds 42/43/44
& Training seed; Student-$t$ ($df=2$)
& Replicated descriptive direct-provenance result. \\

Figure~\ref{fig:ledger-downstream} source suppression
& Persistent and latest-write-only, seeds 42/43/44
& Training seed for cross-checkpoint summaries; paired sequence bootstrap within checkpoint
& Replicated downstream intervention result. \\

SA-MoE exhaustive ranking diagnostic
& Persistent seed 42 per family
& 24 source sequences; 2,000 paired sequence-bootstrap resamples
& Fixed-checkpoint forensic comparison; not a three-seed architecture claim. \\

Ordinary-MoE Li attribution
& Ordinary MoE seeds 42/43/44
& Training seed for cross-checkpoint summaries; sequence bootstrap within checkpoint
& Replicated post-hoc attribution comparison on a distinct audit object. \\

Ledger numerical reconstruction
& Frozen construction checkpoints
& Maximum deterministic numerical residual
& Implementation/numerical validation, not a statistical claim. \\

State-geometry diagnostic
& Older fixed SmolLM2 checkpoint where stated
& Descriptive
& Mechanism/control diagnostic only. \\

Audit-cost pilot
& One frozen workload per family on common hardware
& Descriptive repeated timing phases
& Operational diagnostic; not a training-seed or universal speed claim. \\
\bottomrule
\end{tabular}
\end{table*}

Across all replicated analyses, the paper reports the three individual
training-seed values rather than treating the large number of tokens, probe
examples, routing decisions, or bootstrap draws as independent model
replicates. Same-sign effects across all three seeds are useful replication
evidence, but do not constitute a separate high-powered significance test.

\end{document}